\documentclass[lettersize,journal]{IEEEtran}
\usepackage{amsmath,amsfonts}
\usepackage{algorithmic}
\usepackage{array}
\usepackage[caption=false,font=normalsize,labelfont=sf,textfont=sf]{subfig}
\usepackage{textcomp}
\usepackage{stfloats}
\usepackage{url}
\usepackage{verbatim}
\usepackage{graphicx}
\usepackage{xcolor}
\usepackage{soul}
\usepackage{booktabs}
\usepackage{multirow}

\def\BibTeX{{\rm B\kern-.05em{\sc i\kern-.025em b}\kern-.08em
    T\kern-.1667em\lower.7ex\hbox{E}\kern-.125emX}}
\usepackage{balance}
\begin{document}
\title{An Interactive Evolutionary Multiobjective Optimization Algorithm Based on Interpretable Active Preference Learning}

\author{Thomas Quadt*, Boris Čule, Roy Lindelauf, Herman Monsuur, Mark Voskuijl
\thanks{* Corresponding author: t.j.quadt@tilburguniversity.edu}
\thanks{Thomas Quadt, Boris Čule, and Roy Lindelauf are with the Department of Intelligent Systems, Tilburg University, Tilburg, NL}
\thanks{Thomas Quadt, Roy Lindelauf, Herman Monsuur, and Mark Voskuijl are with the Data Science Center of Excellence, Faculty of Military Sciences, Breda, NL}
}
% \markboth{IEEE Transactions on Evolutionary Computation}{}

\maketitle

\begin{abstract}
Many decision-making tasks are inherently multi-objective. Interactive evolutionary multi-objective optimization algorithms (iEMOAs) are particularly well-suited to help the decision-maker find their most preferred solution. However, many of these algorithms employ black-box preference models. For high-stakes decision-making tasks, like healthcare and defense, a high level of transparency is needed, for instance to support accountability. We propose an interpretable iEMOA which uses active learning to efficiently learn a tree-based preference model directly from pairwise comparisons. The survival operator of NSGA-II and NSGA-III is extended with the learned preference model to guide the algorithm towards the region of interest. An extensive experimental study compares the proposed method to a state-of-the-art black-box model and a state-of-the-art interpretable model. Results indicate that our proposed model outperforms the alternatives in terms of finding the most preferred solution at the cost of a slightly worse average final population. Finally, a case study on a real-life engineering problem illustrates how inspecting the learned model reveals which objectives drive the decision-maker's preferences.
\end{abstract}


\section{Introduction}
\IEEEPARstart{M}{any} real-life optimization tasks are Multi-objective Optimization Problems (MOPs) involving multiple conflicting objectives. For example, when choosing an investment portfolio, one wants to maximize return and minimize risk. 
% Similarly, the best bridge design should minimize cost whilst maximizing structural reliability. 
For these problems, there is not a single optimal solution but rather a set of Pareto optimal solutions called the Pareto front. Whilst mathematically speaking all Pareto optimal solutions are equally good, a Decision-Maker (DM) usually prefers some optimal solutions over others. One example is pandemic policy design during COVID-19, where policy-makers needed to simultaneously minimize mortality, economic damage, and social and psychological harm, such as mental health deterioration. These objectives are inherently conflicting, leading to multiple different optimal policies and thus different policy choices across countries. For instance, Sweden adopted relatively less restrictive measures, aiming to limit societal disruption, whereas the Netherlands implemented stricter interventions, including lockdowns and a curfew, placing greater emphasis on controlling infection rates and protecting healthcare capacity.

% A structural engineer would not choose the cheapest bridge design if the structural reliability is poor. Furthermore, improving the structural reliability by $0.1\%$ at the cost of hundreds of millions might also not be desired.

Evolutionary Multi-objective Optimization Algorithms (EMOAs) are particularly well-suited for MOPs because they maintain a population of solutions, allowing them to approximate multiple trade-off solutions along the Pareto front in a single run. Usually these algorithms first approximate the entire Pareto front after which a DM has to choose their most preferred solution. However, this is computationally costly and when there is a large set of alternatives choosing the most preferred solution is difficult~\cite{Miettinen2008}. Moreover, EMOAs struggle when the number of objectives increases because nearly all solutions become Pareto optimal~\cite{branke2014learning, li2015many}. Alternatively, instead of choosing the most preferred solution after the optimization process, one could introduce a DM's preference information prior to the optimization. This preference information is used to identify a region of interest (ROI), which reduces the search space, and creates an additional ordering in non-dominated solutions increasing the selection pressure. Yet, a DM's limited knowledge of the problem prior to the optimization can lead to inaccurate and misleading preferences~\cite{wang2017mini, thiele2009preference}. Interactive methods, on the other hand, alternate optimization with preference elicitation, allowing the DM to get a better understanding of the problem, the objective trade-offs and the feasibility of their preferences which in turn enables them to make a better-informed choice~\cite{Belton2008, Afsar2021, li2020does}.

These interactive EMOAs (iEMOAs) ask a DM to elicit some preference information, build a preference model, and adapt the optimization algorithm to use this preference information to converge its population towards the ROI. When building such a preference model, one has to consider that a DM's time is costly. Additionally, answering questions needs to be cognitively simple. Direct questions, like numerically rating solutions, are too cognitively demanding~\cite{larichev1992cognitive, GRECO2008416}. On the other hand, indirect preference elicitation builds a preference model via easy-to-answer holistic questions. One of the most common approaches is to ask a DM to answer pairwise comparisons. Stating that one prefers solution A over solution B is namely much easier than numerically rating both, especially when needing to be consistent over a large sets of solutions. To not cognitively overload a DM, it is thus important to learn a preference model in as few easy-to-provide (i.e. indirect) preference statements as possible.

% Interactive multi-objective optimization algorithms can be classified according to their interaction pattern, the type of preference information that the DM needs to supply, the preference model that is formed, and the optimization algorithm that uses the preference model~\cite{Xin2018}.  (indirect preference information, dominance-based EA,  )

Recently, Misitano et al.~\cite{misitano2022towards} have argued for an additional desirable property: explainability. They argue that most current interactive methods can be seen as black-box models. DMs elicit their preferences and obtain a solution, however, they have no idea how their preferences have led to that solution. Yet in critical decision-making tasks it is vital that the decision to be made is well-understood, trusted, and justified. Misitano~\cite{misitano2024enhancing} even questions whether relying on black-box models in any form of decision-making is "good" decision-making at all. Besides context requiring transparent models, providing explanations to DMs can help them better articulate and refine their preferences, speeding up the convergence to the most preferred solutions~\cite{gutierrez2026, rodemann2026explaining, Chakraborty2026}. Although this need for explainability in AI has long been stressed, explainability in iEMOAs is still a very new research field~\cite{osika2023lies, Misitano2025} and only a few (post-hoc) explainable or (inherently) interpretable methods exist.

% Consequently, there is a tension between the effectiveness of current iEMOAs and the transparency required for trustworthy decision-making. While preference elicitation can steer the search, the absence of interpretable preference models leaves decision makers unable to understand why certain solutions are obtained and others are not.

We propose a novel inherently interpretable iEMOA called DT-PBO-NSGA. Our proposed method uses active learning to build a semi-Bayesian decision tree as preference model. This tree partitions the objective space in a set of leaves and ranks each leaf with a preference value with the highest ranked leaf representing the DM's ROI. By using the decision tree's learned ranking within NSGA-II~\cite{Deb2002} or NSGA-III's~\cite{nsga3} survival operator, we direct the population to converge towards this ROI. Experimental results on an extensive suit of synthetic benchmarks show that our model performs similar to state-of-the-art baselines (DTEMOA~\cite{Shavarani2023} and PBO-R-NSGA-III~\cite{Taylor2021}) in lower dimensions and is more conservative in reducing its ROI, leading to the ROI more often containing the most preferred solutions whilst having a slightly worse average final population. Moreover, a case study on real-world engineering problem of the crashworthiness of car designs shows the learned preference model's inherent interpretability.

The remainder of this paper is organized as follows. In Section~\ref{sec:related_work}, we review a set of iEMOAs in terms of some key characteristics.
% : the preference model they employ, the preference information they request from the DM, the type of evolutionary algorithm they extend, and the way they determine what preference statement to ask next. 
The review motivates the development of our proposed methodology. In Section~\ref{sec:preference_model}, we give a short overview of our interpretable preference model based on interpretable Preferential Bayesian Optimization~\cite{leenders2026}. In Section~\ref{sec:iPBO_NSGA-II} we describe how to integrate the preference model with the well-known NSGA-II~\cite{Deb2002} and NSGA-III~\cite{nsga3} algorithms. In Section~\ref{sec:numerical_experiments}, we discuss the results of our numerical experiments on synthetic benchmarks and in Section~\ref{sec:case_study} we apply DT-PBO-NSGA to a real-world engineering problem showing its preference model's inherent interpretability. Finally, in Section~\ref{sec:conclusion}, we provide our conclusions and introduce interesting directions for future research.

\section{Related Work}\label{sec:related_work}
Interactive EMOAs differ in the following aspects: the type of optimization algorithm they adapt, the preference model they employ, the preference information they request from the DM, and the way they determine what preference statement to ask next. A summary of these key features for a selection of 20 research studies is shown in Table~\ref{tab:LitReview}. The papers that are selected are a representative subset of interactive EMOAs that utilize indirect preference judgments. This overview of papers is not meant to be a comprehensive survey, but instead shows the main research developments of interactive EMOAs in the aforementioned aspects over the years. For more comprehensive surveys we refer the readers to: Xin et al.~\cite{Xin2018interactive}, Larraga et al.~\cite{Larraga2026}, and Ren et al.~\cite{Ren2026}.

\begin{table*}
\centering
\caption{Summary of Interactive Evolutionary Multi-objective Algorithms utilizing indirect preference judgments. (UF = Utility Function, PCs = Pairwise Comparisons *-FGKKMW are the initials of the author and PBO-R-NSGA-III is not an official name but a descriptive label reflecting the approach).}
\label{tab:LitReview}
\begin{tabular}{@{}lllllll@{}}
\toprule
\textbf{Algorithm} & \textbf{Year} & \textbf{EA Type}    & \textbf{Preference Model}  &\textbf{Interpretable}& \textbf{Preference Information} & \textbf{Question Selection} \\ \midrule
IEM~\cite{Phelps2003} & 2003          & Dominance     & Linear Weighted Sum UF     &High& PCs            & Balanced                             \\
PI-NSGA-II-UF~\cite{Deb2010}      & 2010          & Dominance     & Polynomial UF              &Medium& Ranking                         & Diverse Set                          \\
BC-EMOA~\cite{Battiti2010}            & 2010          & Dominance     & Support Vector Machine UF  &Low& Ranking/PCs    & Greedy                \\
FGKKMW*~\cite{Fowler2010}            & 2010          & Dominance     & Quasi-concave UF           &Low& Polarized Solution              & Random                               \\
iTDEA~\cite{Koksalan2010}              & 2010          & Dominance     & Preference Region UF       &Medium& Polarized Solution              & Greedy                \\
iMOEA/D~\cite{Gong2011}            & 2011          & Decomposition & Preference Region UF&Medium& Polarized Solution              & Random                               \\
W-HypE~\cite{Brockhoff2014}             & 2014          & Indicator     & Weighted Hypervolume UF    &Medium& Polarized Solution              & Not Specified                        \\
NEMO-0,I,II~\cite{Branke2015}        & 2015          & Dominance     & Set of Additive Monotonic UFs&Medium& PCs            & Random                               \\
I-SIBEA~\cite{Chugh2015}            & 2015          & Indicator     & Weighted Hypervolume UF    &Medium& Good vs Bad                     & Greedy                \\
NEMO-II-Ch~\cite{Branke2016}         & 2016          & Dominance     & Choquet Integral UF        &Medium& PCs            & Random                               \\
IEMO/I~\cite{Tomczyk2019}             & 2019          & Indicator     & Set of Chebyshev UFs&Medium& PCs            & Random                               \\
INFRINGER~\cite{Misitano2020}          & 2020          & Not mentioned       & Belief Rule UF             &High& PCs            & Diverse Set                          \\
IEMO/D~\cite{Tomczyk2020}             & 2020          & Decomposition & Set of $L_{\alpha}\text{-norm}$ UFs&Medium& PCs            & Random                               \\
PBO-R-NSGA-III*~\cite{Taylor2021}     & 2021          & Decomposition & Gaussian process UF        &Low& PCs            & Active Learning                      \\
iDMOEA-$\epsilon$C~\cite{Chen2021} & 2021          & Decomposition & Preference Region          &Medium& Polarized Solution              & Diverse Set                          \\
DT-EMOA~\cite{Shavarani2023}            & 2023          & Dominance     & Decision Tree UF           &High& Ranking            & Random                               \\
I-EMO/LTR~\cite{Li2023}          & 2023          & All                 & Neural Network UF          &Low& PCs            & Random                               \\
XIMEA-DRSA~\cite{Corrente2024}         & 2024          & Dominance     & Decision Rules             &High& Good vs Bad                     & Greedy                \\
 D-PBEMO~\cite{huang2024}& 2024& Dom. + Decomp.& Dueling Bandits& Medium& PCs&Active Learning\\
RF-EMOA~\cite{Shavarani2025}           & 2025          & Dominance     & Random Forest UF           &Medium& PCs            & Active Learning                      \\ 
ERS~\cite{tomczyk2026efficient} & 2026 & Not Mentioned& Set of $L_{\alpha}\text{-norm}$ UFs&Medium& PCs & Not Mentioned\\ \bottomrule
\end{tabular}
\end{table*}

\subsection{Type of Evolutionary Algorithm}
EMOAs can be largely classified in three classes: dominance-based, decomposition-based, and indicator-based. Many interactive EMOAs are based on dominance-based algorithms. These algorithms use the Pareto-dominance concept to select solutions for survival. In higher dimensions, however, nearly all solutions become non-dominated which severely degrades the selection pressure of the algorithm~\cite{ishibuchi2008evolutionary}. Incorporating preference information into the survival operator can reintroduce this selection pressure by focusing on a small preferred region of the Pareto front~\cite{Branke2015}. Recently, more attention has gone towards extending decomposition-based EMOAs with preference information. These methods decompose the problem into a set of single-objective optimization problems, usually via weights representing an instance of a scalarizing function or via reference directions. Each set of weights generates a different single-objective optimization problem and preference information is used to steer the set of weights to represent problems in the preferred regions. Lastly, indicator-based EMOAs use a performance indicator (e.g. hypervolume) to measure the performance of a set of solutions. The survival operator is then fully based on this indicator; the set of solutions that maximizes the indicator is chosen to survive. Only a few interactive indicator-based EMOAs exist, but generally they employ a weighted version of the indicator and adapt the weights to adhere to these preferences. 
Nearly all methods extend a single type of EMOA with preference information. The only exception is I-EMO/LTR\cite{Li2023}, where they created a preference-based version for all three types.

\subsection{Preference Model}
Most preference models are based on latent utility functions defined over the objective space $\mathcal{F}$. A value or utility function $u(\mathbf{f}):\mathcal{F} \rightarrow \mathbb{R}$ represents a DM's preferences over an objective vector $\mathbf{f}$, where a higher $u(\mathbf{f})$ means a more preferred $\mathbf{f}$. In particular, $\mathbf{f} \succ \mathbf{f'}$ implies $u(\mathbf{f}) > u(\mathbf{f'})$. Over the years, many different utility functions ranging from linear weighted sums~\cite{Phelps2003} to polynomials~\cite{Deb2010} to general quasi-concave functions~\cite{Tomczyk2020} have been used. Each utility function is parametrized by a set of weights that are learned from preference statements supplied by the DM. The functional form of these utility functions has become more complicated, enabling more general preferences to be learned. However, this expressiveness comes at the cost of interpretability. Even in a simple linear weighted sum the interpretation of weights is nontrivial~\cite{Marler2010}, let alone in complex polynomials or quasi-concave functions.

More recently, the use of inherently interpretable models as utility functions has been proposed. First, Misitano et al.~\cite{Misitano2020} proposed INFRINGER, which learns a set of belief rules that aim to represent what entails a good solution. These rules and their associated belief values are then aggregated to get a utility function. The interpretability of the model is not discussed further. Similarly, XIMEA-DRSA~\cite{Corrente2024} learns decision rules that classify solutions as either good or bad, but restricting all solutions to only two classes limits the expressiveness of the preference model. In DT-EMOA~\cite{Shavarani2023}, a decision tree is used to represent the utility function. The proposed classification tree predicts when one solution is preferred over another which allows for local interpretability. In particular, one can see when one item is better than another, but global interpretability, seeing what items are generally the best or the worst is not directly evident. Moreover, the decision tree is learned on the difference in feature values which implicitly assumes that there is a constant tradeoff between objectives over their entire range. However, human preferences tend to be non-linear~\cite{Piegat2012}, a difference in feature vectors at the lower hand is not necessarily as desired as the same difference in high ranges. In RF-EMOA~\cite{Shavarani2025}, the authors replace the decision tree with a random forest ensemble to support active learning, but using an ensemble model comes at the cost of interpretability.

\subsection{Preference Information}
The requested preference information can be either direct or indirect. As direct preference information has been shown to be difficult to supply, the review of papers has solely focused on methods that use indirect preference information. The review shows that most of the methods use pairwise comparisons or polarized solutions (e.g., selecting the best solution out of a set).

\subsection{Question Selection}
Since a DM's time and effort is costly, a preference model needs to be learned in as few preference statements as possible. Choosing the next question to ask in a smart way can help reduce the information needed to learn a model. Usually, it is a good idea to present preference options with a high probability of being the most preferred. However, when only greedily asking questions involving the current best solution, one runs into the risk of ending up in a local optimum. Therefore, active learning methods balance exploration with exploitation. To do so, these methods require a probabilistic estimate of the utility function to quantify a degree of uncertainty. Table~\ref{tab:LitReview} show that only RF-EMOA and PBO-R-NSGA-III use active learning. The rest of the reviewed methods still use random question selection, simply select a diverse representation of the current population, or always greedily ask the current best solution.

\section{Method}
The reviewed literature reveals that whilst interpretable iEMOAs have become more common, there are still two important limitations. First, interpretable preference models are often limited in expressiveness or provide only local interpretability. Second, although pairwise comparisons are commonly used for preference elicitation, Table~\ref{tab:LitReview} shows that most approaches select queries using random or greedy strategies rather than active learning. 
To address these limitations, we extend NSGA-II and NSGA-III with a globally interpretable regression tree preference model that supports active preference elicitation. The model is globally interpretable in the sense that the complete learned preference structure is represented by a single inspectable tree, allowing the decision maker to examine how preferences vary across the objective space rather than being limited to local explanations of individual predictions. 

We use NSGA-II for bi-objective problems and NSGA-III for problems with three or more objectives. For the remainder of this paper we will use NSGA to refer to the applicable variant. We adopt NSGA as the underlying evolutionary algorithms because their front-based selection mechanism naturally aligns with the proposed tree-based preference model. In particular, the preference model partitions the objective space into regions with different predicted preference values, from which preference fronts can be constructed. These fronts can be incorporated directly into NSGA's existing front-based selection procedure, allowing preference information to steer the search without requiring assumptions about the functional form of the DM's preferences.
% We use NSGA as base evolutionary algorithm as indicator-based algorithms have generally shown poor performance and decomposition-based algorithms require a scalarizing function decomposing the problem into single objective optimization problems. As our tree-based method does not assume a parametric form it lends itself better to NSGA which does not need such a scalarizing function. The novel preference model learns a probabilistic estimate of the preference value for each solution directly from pairwise comparisons provided by the decision maker. In particular, each leaf of the tree represents a distribution over preference values, which is used both to rank solutions and to support active preference elicitation. By partitioning the objective space into leaves, the preference model induces preference fronts that steer NSGA's evolutionary search toward preferred regions of the Pareto front. 
The following subsections describe how the preference model is learned from pairwise comparisons, how active learning is used to select informative comparisons, and how the resulting preference model is integrated into NSGA-II and NSGA-III.

\subsection{Preference Model}\label{sec:preference_model}
To model the DM's preferences, we use a recently proposed semi-Bayesian regression tree~\cite{leenders2026}. Learning this model from pairwise comparisons requires partitioning the objective space into leaves and find the corresponding leaf estimates. The leaf estimates are represented probabilistically to capture uncertainty and facilitate active learning via Preferential Bayesian Optimization (PBO). 

\begin{figure}
    \centering
    \includegraphics[width=\linewidth]{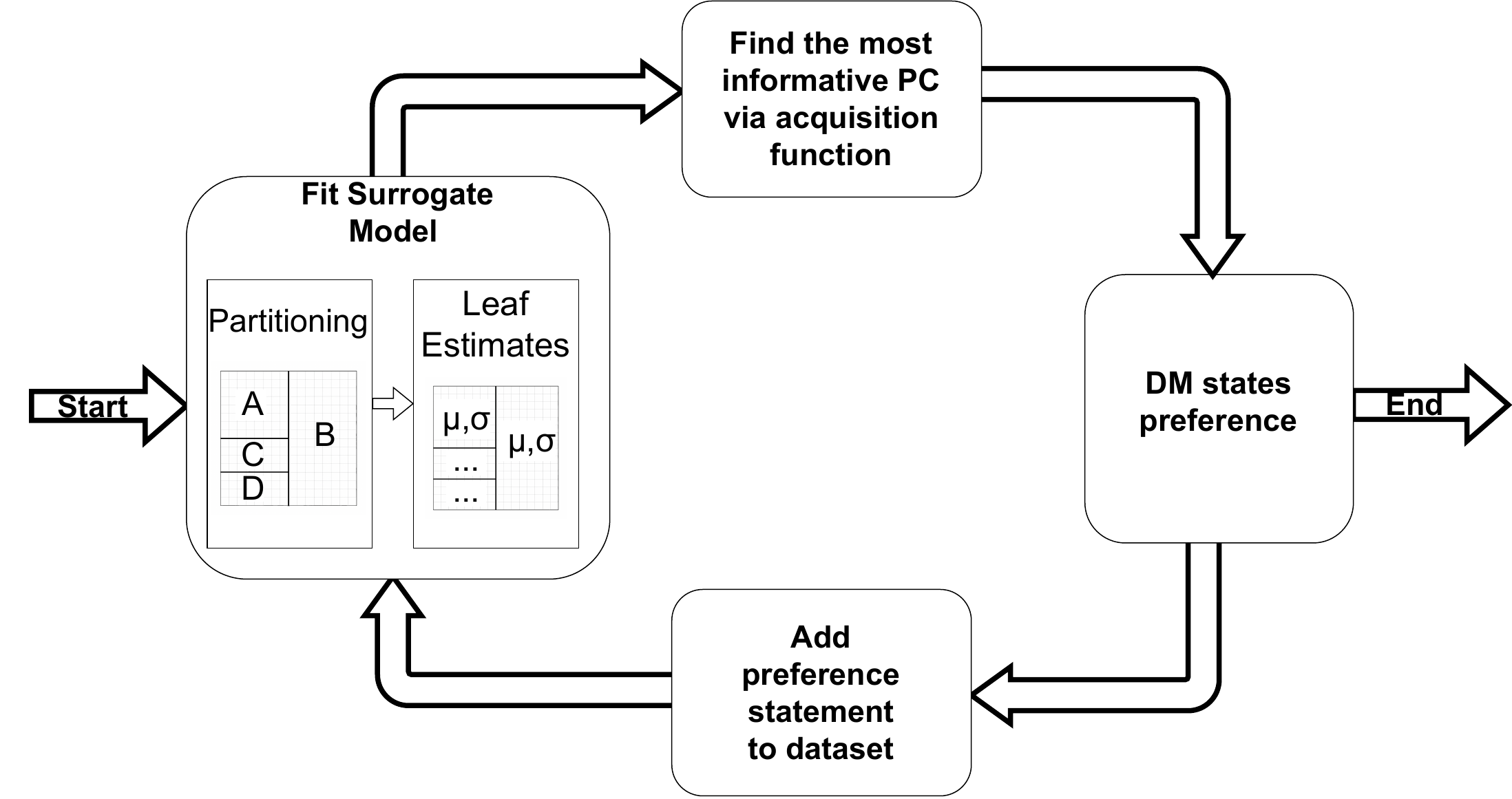}
    \caption{Preferential Bayesian Optimization process of learning a decision tree as preference model. Taken from~\cite{leenders2026}.}
    \label{fig:PBO_process}
\end{figure}

The PBO process is illustrated in Figure~\ref{fig:PBO_process}. Starting from an initial set of pairwise comparisons, a regression tree is learned to model the DM's preferences. Based on this model and its uncertainty, an acquisition function selects the next most informative pairwise comparison. The selected comparison is presented to the DM, whose response is added to the dataset. The regression tree is subsequently updated using the augmented dataset, after which the process is repeated until a stopping criterion is met. This process can continue until a convergence criteria is reached. The following sections give a brief overview of the three main components: (1) the partitioning of the objective space, (2) finding the probabilistic leaf estimates, and (3) the acquisition function used to determine the next most informative question to ask. For a more comprehensive explanation, we refer the reader to the original paper~\cite{leenders2026}. Finally, a working example to illustrate the algorithm's workings is provided.

\subsubsection{Partitioning the Objective Space}
Regular regression trees use a splitting heuristic to greedily partition the input space into leaves. These splitting heuristics use the numerical labels of the training dataset to measure node impurity (e.g., variance) and try to create nodes that are as pure as possible. In our setting we do not have access to numerical labels. Instead, given a tree node $A$, we have a set of $N_A$ pairwise comparisons $D_A = \{(\mathbf{x}_i \succ \mathbf{x'}_i)\}_{i=1}^{N_A}$, where $\mathbf{x} \succ \mathbf{x'}$ means $\mathbf{x}$ is preferred to $\mathbf{x'}$. To determine the purity of a node, a custom splitting heuristic specifically designed for pairwise comparisons is used. The main idea is that the score measures how well preferred items are separated from non-preferred items. A perfect split would namely put all preferred items on one side and the non-preferred items on the other.

A split is defined by a feature index $k$ (where $x_i[j], x_i'[j]$ are the $j^{\text{th}}$ feature value of PC $\mathbf{x}_i \succ \mathbf{x}_i'$) and a threshold value $t$. For a candidate split $s=(k,t)$, define
\begin{equation}\label{eq:Zi_def}
Z_i(s)
=
\mathbf{1}\{x_i[k]\ge t,\;x_i'[k]<t\}
-
\mathbf{1}\{x_i[k]<t,\;x_i'[k]\ge t\},
\end{equation}
so that $Z_i(s)\in\{-1,0,1\}$: $+1$ has the winner on the right and the loser on the left of the split, $-1$ the reverse, and $0$ has both items on the same side.
The consistency score is then

\begin{equation}\label{eq:consistency}
\widehat S_A(s)
=
\left|
\frac{1}{N_A}
\sum_{i=1}^{N_A} Z_i(s)
\right|.
\end{equation}

A high value of $\widehat S_A(s)$ means that node $A$ clearly separates the winners from the losers. 

The tree is constructed greedily by selecting the $k*$ and $t*$ that maximize the Consistency score. After a split, the child node datasets $D_{\textit{left}}$ and $D_{\textit{right}}$ are formed only by pairs that fall entirely on one side of the partition:
\begin{equation}\label{eq:split_left_right}
\begin{aligned}
    D_{\textit{left}} &= \{(\mathbf{x}_i, \mathbf{x'}_i) \in D \mid x_i[k^*] < t^* \land x_i'[k^*] < t^*\} \\
    D_{\textit{right}} &= \{(\mathbf{x}_i, \mathbf{x'}_i) \in D \mid x_i[k^*] \geq t^* \land x_i'[k^*] \geq t^*\}
\end{aligned}
\end{equation}
Pairs that straddle the split (and can thus already be differentiated by the decision tree) are discarded. By discarding these points, the algorithm reduces the internal dataset size which in turn reduces the size of the final decision tree leading to shallower more interpretable trees. At the same time this self-controlling growth removes the need for pre- and post pruning, although one could still set the maximum depth of the tree, minimal split score and minimal samples per split manually.

\subsubsection{Finding the Probabilistic Leaf Estimates}
After partitioning the objective space into $m_l$ leaves, the probabilistic estimates of the leaf values are obtained using Bayes' Rule:

\begin{equation*}
    p(\mathbf{f}|\mathcal{D}) \propto p(\mathcal{D}|\mathbf{f})p(\mathbf{f}).
\end{equation*}
For the prior $p(\mathbf{f})$, it is assumed that each leaf follows an independent zero-mean Gaussian with variance $\sigma_{prior}^2$, which leads to the following joint multivariate Gaussian
\begin{equation}\label{eq:leaf_prior}
    \mathbf{f} \sim \prod_{j=1}^{m_l} \mathcal{N}(0,\sigma_{prior}^2) = \mathcal{N}(\mathbf{0}, \sigma_{prior}^2I), 
\end{equation}

$\text{where } I \text{ is the $m_l$-dimensional identity matrix}.$
For the likelihood $p(\mathcal{D}|\mathbf{f})$, it is assumed that observations are influenced by homoskedastic Gaussian noise. Assuming independence of preference statements the following likelihood function~\cite{Benavoli2026} is obtained:

\begin{equation}\label{eq:likelihood}
    p(\mathcal{D}_n|\mathbf{f}) = \prod_i^{n}\Phi(\frac{f(\mathbf{x}_i) - f(\mathbf{x}'_i)}{\sqrt{2}\sigma_{noise}}),
\end{equation}
 where $\Phi(z) = \int_{-\infty}^z \mathcal{N}(\mathbf{x};0, 1)d\mathbf{x}$ is the cumulative standard normal distribution.

The resulting posterior distribution is computationally intractable. Therefore, Laplace Approximation is used to approximate the distribution with a Gaussian centered at the mode (i.e., maximum a posteriori (MAP) estimate) of the posterior. In particular, $\mathbf{f}|\mathcal{D} \sim \mathcal{N}(\mathbf{f}_{MAP}, \Sigma_{post})$ where $\mathbf{f}_{MAP} = \arg \max_{\mathbf{f}}p(\mathbf{f}|D)$, which can be found as the minimizer of the following functional

\begin{equation}\label{eq:functional}
    L(\mathbf{f}) = - \sum_{i=1}^n \ln \Phi(\frac{f(x_i) - f(x'_i)}{\sqrt{2}\sigma_{noise} }) + \frac{1}{2\sigma_{prior}^2}\mathbf{f}^\top \mathbf{f}.
\end{equation}

The posterior Covariance is defined as
\begin{equation}\label{eq:DT_cov}
    \Sigma_{post}^{-1} = H(L(\mathbf{f}))|_{\mathbf{f} = \mathbf{f}_{MAP}} = ( \frac{1}{\sigma_{prior}^2}I + \Lambda_{MAP}).
\end{equation}
where $\Lambda$ is an $m \times m$ matrix where the $ij$th element is $\frac{\partial^2 \sum_{k=1}^n - \ln \Phi(z_k)}{\partial f(\mathbf{x}_i)\partial f(\mathbf{x}_j)}$, $z_k = \frac{f_{MAP}(\mathbf{x}_k) - f_{MAP}(\mathbf{x}'_k)}{\sqrt{2}\sigma_{noise} }$.\\

Note that the likelihood is translationally invariant with respect to $\mathbf{f}$ (i.e. $p(\mathcal{D}|\mathbf{f}) = p(\mathcal{D}|\mathbf{f} + c \mathbb{I})$, where $c \in \mathbb{R}$, and $\mathbb{I}$ is the indicator matrix). Adding a constant shift $c$ to $\mathbf{f}$ does not influence the pairwise differences and thus leaves the likelihood unchanged. This translational invariance introduces a large irreducible variance component in the constant shift direction. To remedy this all leaves are constrained to sum up to zero ($\mathbf{1}\mathbf{f} = 0$), which fixes the scale of the model and prevents a constant shift. This constraint is incorporated by using Eaton's \cite{Eaton2007} normal conditional distribution which gives the following conditional distribution

\begin{equation}\label{eq:conditional_distribution}
\mathbf{f}\;\bigl|\;\mathbf{1}^\top\mathbf{f} = 0 \;\sim\; \mathcal{N}\!\Bigl(\mathbf{f}_{MAP}, \Sigma_{\mathrm{post}}-\frac{\sigma_{\mathrm{prior}}^2}{m_l}\mathbf{1}\mathbf{1}^{\top}\Bigr)
\end{equation}

The leaf-fitting procedure contains two hyperparameters $\sigma_{noise}$ and $\sigma_{prior}$, however, as shown by Leenders et al.~\cite{leenders2026} the hyperparameters are not identifiable separately. Therefore, the fitting procedure only depends on the ratio of the two: $\sigma_{{noise}}^2 / \sigma_{{prior}}^2$. Increasing the ratio essentially moves the utility values towards zero, whilst decreasing the ratio moves them farther from zero. However, for non-extreme ratios, the effect on performance is minimal, as the ratio mainly affects the scale of the utilities rather than the relative ordering of the leaves.

\subsubsection{Active learning}
The previous two subsections explained how a decision tree with probabilistic leaves can be learned from pairwise comparison data. However, to learn a preference model as efficiently as possible, one also needs to determine which question to ask. Currently, the state-of-the-art acquisition function used in active preference learning is qEUBO~\cite{astudillo2023qeubo}. However, qEUBO is originally defined for continuous models (Gaussian processes). Instead, the learned regression tree is a discrete representation. Two points falling in the same leaf cannot be differentiated from one another. qEUBO will never select these points. Therefore, a structural refinement term $I_{\text{struct}}$, which represents the expected change in consistency score within the leaf containing the pair, is added. For a candidate pair $(\mathbf{x},\mathbf{x}')$ falling in leaves $A_i$ and $A_j$, qEUBO is changed as follows:

\begin{equation}\label{eq:AF}
    \alpha_{\text{DT}}(\mathbf{x},\mathbf{x}') =
\text{qEUBO}(\mathbf{x},\mathbf{x}') +
\lambda \cdot I_{\text{struct}}(\mathbf{x},\mathbf{x}'),
\end{equation}
where $\lambda \geq 0$ controls how strongly the acquisition prioritizes structural refinement relative to qEUBO and $I_{\text{struct}}$ is defined as

\begin{equation}
    I_{\text{struct}}(\mathbf{x},\mathbf{x}') = \mathbb{E}_Y\!\left[\big|\widehat{S}_A^{(+Y)}(s_A^\star) - \widehat{S}_A(s_A^\star)\big|\right]\cdot\mathbf{1}\{A_i = A_j = A\},
\end{equation}
where $Y\in\{\mathbf{x}\succ\mathbf{x}',\mathbf{x}'\succ\mathbf{x}\}$ is the unobserved outcome, $\widehat{S}_A^{(+Y)}$ is the consistency score recomputed after adding the comparison with outcome $Y$ to $D_A$, and the expectation is taken under the Bernoulli distribution. Within a leaf, $P(\mathbf{x}\succ\mathbf{x}')=\Phi(0)=0.5$, so the expectation reduces to the average over the two outcomes. The indicator term ensures that the structural term is only active for two items falling in the same leaf. Note that pairwise comparisons between leaves can be seen as a form of exploration, whereas pairwise comparisons within a single leaf are aimed at exploitation. Therefore, within-leaf comparisons are only done on the current best-leaf.

 After choosing which leaves to use for generating the next most informative question, the actual points in the leaves still need to be selected. These points are chosen at random. Specifically for our interactive optimization procedure, points falling in the chosen leaves which are also in the current population or external archive (defined in Section~\ref{sec:iPBO_NSGA-II}) can be sampled.  

\subsubsection{Working Example}
\begin{figure}
    \centering
    \includegraphics[width=\linewidth]{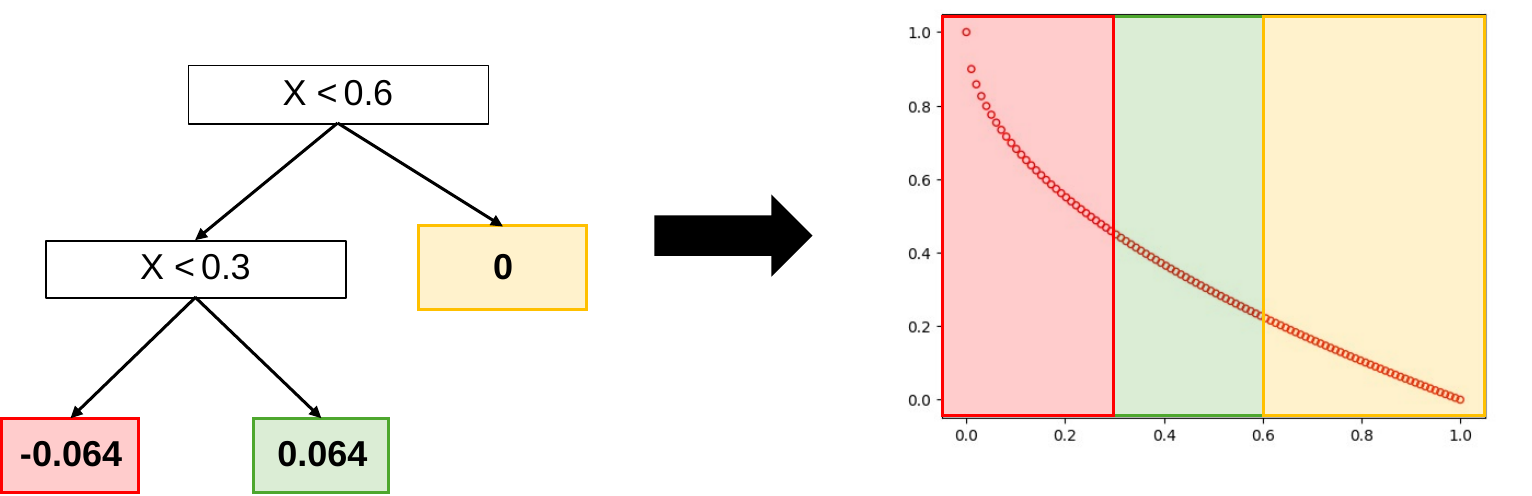}
    \caption{Decision tree learned from pairwise comparisons $\{\mathbf{x}_{mid} \succ \mathbf{x}_{left}, \mathbf{x}_{mid} \succ \mathbf{x}_{right}, \mathbf{x}_{left} \succ \mathbf{x}_{right}\}$ , where $\mathbf{x}_{left} \approx (0.2, 0.55)$, $\mathbf{x}_{mid} \approx (0.40, 0.37)$, $\mathbf{x}_{right} \approx (0.8, 0.11)$ and its resulting partitioning and ranking of the objective space of ZDT1.}
    \label{fig:exampleDT}
\end{figure}

To illustrate how our preference model is learned, we use the 2-dimensional ZDT-1~\cite{zitzler2000comparison} MOP benchmark as an example. Given three pairwise comparisons as follows: $\{\mathbf{x}_{mid} \succ \mathbf{x}_{left}, \mathbf{x}_{mid} \succ \mathbf{x}_{right}, \mathbf{x}_{left} \succ \mathbf{x}_{right}\}$, where $\mathbf{x}_{left} \approx (0.2, 0.55)$, $\mathbf{x}_{mid} \approx (0.40, 0.37)$, $\mathbf{x}_{right} \approx (0.8, 0.11)$, we first use our consistency score heuristic to determine the structure of the tree and then find the probabilistic leaf estimates. Two splits can be made. The split between $\mathbf{x}_{left}$ and $\mathbf{x}_{mid}$, at $t_x=0.3$, and the split between $\mathbf{x}_{mid}$ and $\mathbf{x}_{right}$ at $t_x=0.6$. The first split at $t_x=0.3$ splits $\mathbf{x}_{mid} \succ \mathbf{x}_{left}$ by putting the winner $\mathbf{x}_{mid}$ on the right, setting $Z_1$ to $1$ and it splits $\mathbf{x}_{left} \succ \mathbf{x}_{right}$, by putting the winner $\mathbf{x}_{left}$ on the left, setting $Z_2$ to $-1$. Computing the consistency score then yields $\widehat S_A(x, 0.3) = |\frac{1}{3} * (1+ 0 - 1)| = 0$. For the second possible split at $t_x=0.6$, $\mathbf{x}_{mid} \succ \mathbf{x}_{right}, \text{ and } \mathbf{x}_{left} \succ \mathbf{x}_{right}$ are both split by the threshold. This split moves two winners to the left, setting $Z_1 = Z_2 = -1$. Therefore, the consistency score is $\frac{2}{3}$ and this split is chosen as the optimal split. In the next iteration, the comparisons that were split are discarded from the data. The only remaining pairwise comparison is $\mathbf{x}_{mid} \succ \mathbf{x}_{left}$. Therefore, the split is done on $t_x = 0.3$. The probabilistic leaf estimates can then be found using Equations~\ref{eq:functional}, \ref{eq:DT_cov}, and \ref{eq:conditional_distribution}. The hyperparameters determine the exact scale of the leaf values, but using arbitrarily chosen $\sigma_{noise}=100, \sigma_{prior}=2$ we get the decision tree and its partitioning in the objective space of ZDT1 as shown in Figure~\ref{fig:exampleDT}. The posterior means of each leaf are visualized as well as the ranking they created. The most preferred region of the objective space as modeled by this tree is between $0.3 \leq x < 0.6$.

After learning the decision tree, we find the most informative pairwise comparison by maximizing $\alpha_{DT}$ as defined in Equation~\ref{eq:AF}. Next, we randomly sample points in the chosen leafs and supply the preference statements to the decision-maker. Finally, the model is updated using the given preference feedback.

\subsection{Integration with Evolutionary Algorithm}\label{sec:iPBO_NSGA-II}
\begin{figure}
    \centering
    \includegraphics[width=\linewidth]{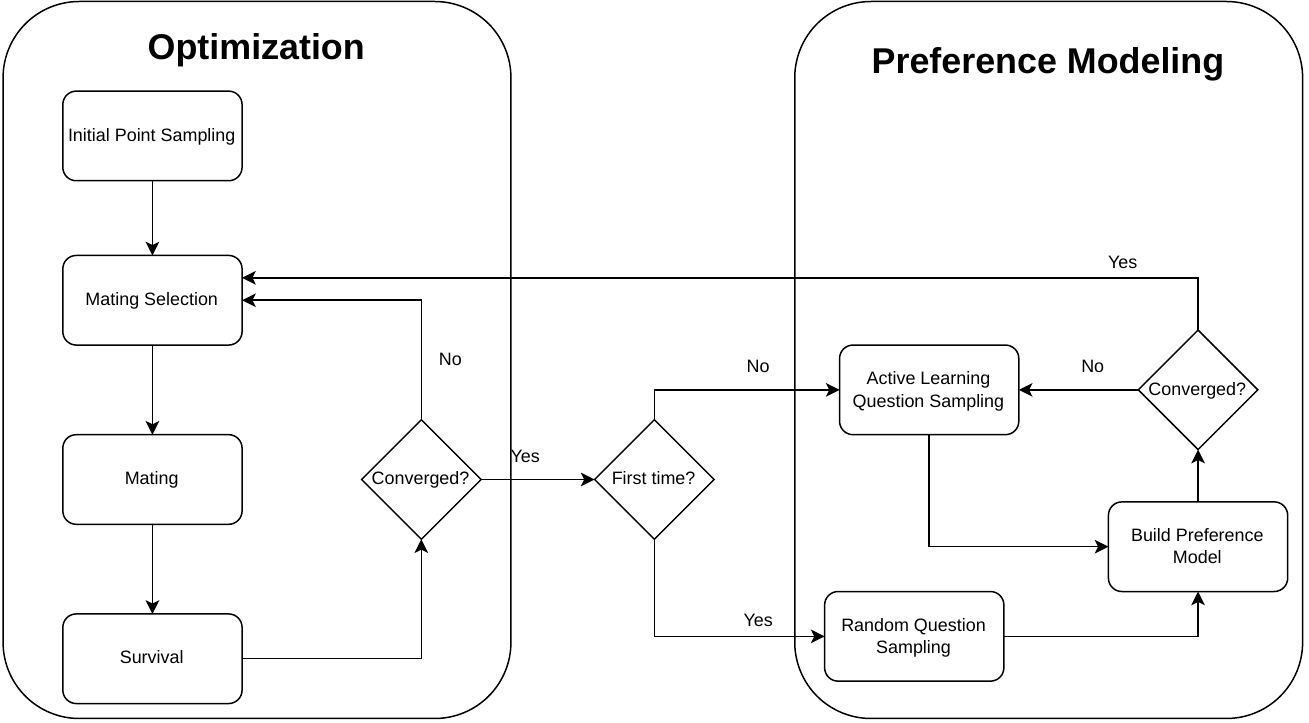}
    \caption{High-level workflow of DT-PBO-NSGA.}
    \label{fig:iNSGA-II}
\end{figure}

Interactive optimization procedures alternate preference elicitation with optimization. Our procedure, as visualized in Figure~\ref{fig:iNSGA-II} starts with running NSGA to get an initial estimate of the Pareto front. After the convergence criteria (e.g., number of iterations) is reached, we switch to preference elicitation. For the first preference modeling stage, we start with randomly asking a few pairwise comparisons. Based on this information the preference model is constructed. For subsequent stages, we ask the most informative questions according to our active learning strategy. If the convergence criteria is reached we switch back to the optimization procedure and continue this process until the decision-maker is satisfied with the current Pareto front. When the optimization procedure is active with a learned preference model, we use this preference information to guide the population to the ROI. To do so, we adapt NSGA's survival operator and its mating selection operator to use the learned preference information. Moreover, we maintain a representative subset of the entire Pareto front (called external archive) to prevent ending up with dominated solutions, accommodate between-leaf comparisons, and quickly adapt to preference changes.

\begin{figure}
    \centering
    \includegraphics[width=\linewidth]{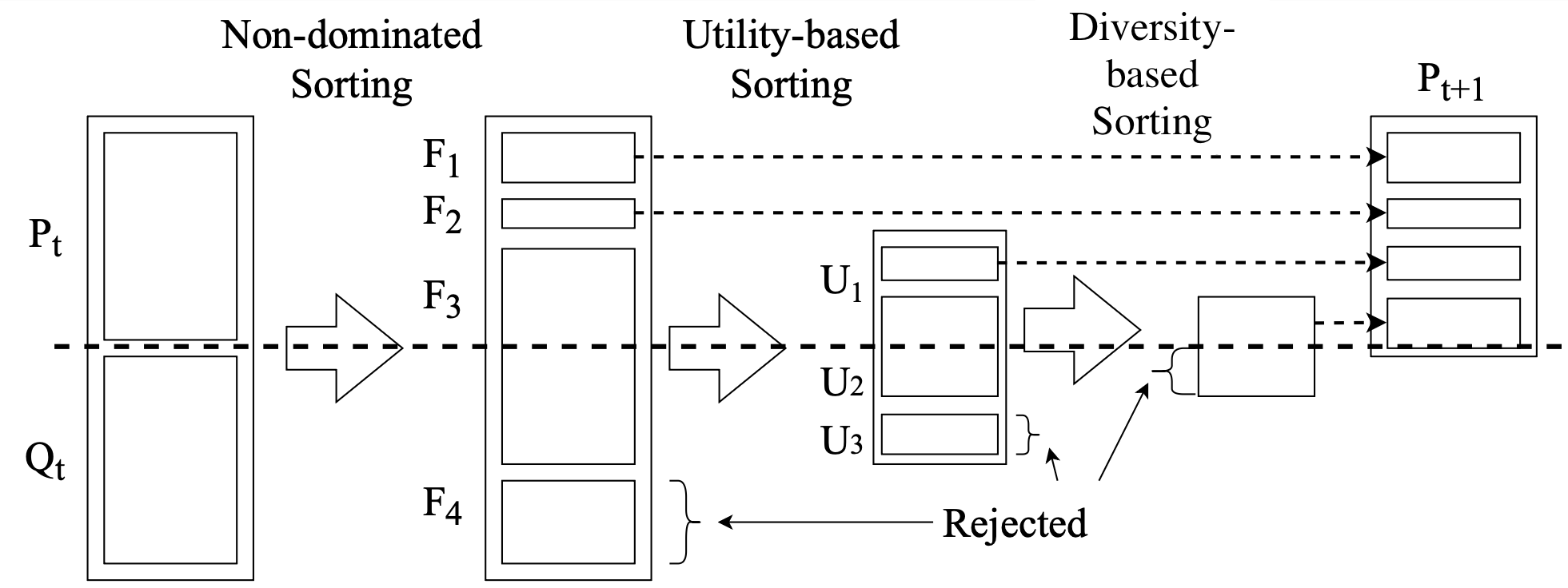}
    \caption{Survival operator of DT-PBO-NSGA. Figure adapted from \cite{Deb2002}.}
    \label{fig:survival_op}
\end{figure}

\paragraph{Survival Operator}
The preference-based NSGA's surival operator is focused on three goals, which in order of importance, are 1) to find Pareto-optimal solutions 2) in the preferred region 3) as diversely spread as possible. To achieve these goals, the survival operator, as visualized in Figure~\ref{fig:survival_op}, uses the non-dominated sorting and diversity-based sorting steps from the original NSGA algorithms, but with a preference-based sorting step in between as follows:

\begin{enumerate}
    \item \textit{Promote Pareto-optimality:} For each solution it is counted by how many solutions it is dominated. The solutions are sorted in domination fronts according to these domination counts. Solutions with the lowest domination counts survive. When a domination front has more individuals than the remaining solutions that are allowed to survive, this front has to be split.

    \item \textit{Promote preferred region:} For each solution in the current domination front the preference value is estimated. The decision tree constructs preference fronts which are used in the same way as the domination fronts. When a preference front needs to be split the next goal is used.

    \item \textit{Promote diversity:} The remaining solutions are sorted according to their diversity-based score. For NSGA-II crowding distance is used, for NSGA-III the reference-direction-based sorting is used. The most diverse points in the current preference region survive.
\end{enumerate}

\paragraph{Mating Selection Operator}
We use binary tournament selection with replacement to select the parents for mating. Similarly to the survival operator, the goals, in order of importance, are Pareto-optimality, Preference, Diversity. Therefore, our operator first checks whether a solution dominates the other. If not, it checks whether one solution is preferred over the other, and finally if that is also not the case, the solution with the better diversity-based score is selected.

\paragraph{External Archive}
Besides changing the survival and mating selection operators, we also maintain an external archive (i.e. a representative subset of non-dominated solutions over the entire Pareto front) for the following reasons: First, due to the interactive optimization process, the current population will converge towards being fully contained in the current best leaf. To ask comparisons of items falling in different leaves we need to store a representative subset of non-dominated solutions over the entire Pareto front. Second, NSGA, especially when including preference information, can end up with dominated solutions when the true Pareto front is disconnected~\cite{dasdemir2020flexible}. Using the external archive in the non-dominated sorting step in the survival operator, can detect these dominated solutions and ensure only non-dominated solutions are returned. Finally, when the preference model shifts towards a new region, for example due to a change of preferences of the DM, the archive can help in quickly determining the new Pareto front. In particular, by merging the current population with the external archive upon a preference shift, we can quickly find solutions close to the new preferred region.

The size of this archive is twice the population size to be able to represent a large diverse Pareto front. It is maintained by merging the current population with the archive after each iteration and using the regular NSGA survival operator to maintain an as diverse as possible non-dominated representation of the entire Pareto front.

\begin{figure}
    \centering
    \includegraphics[width=\linewidth]{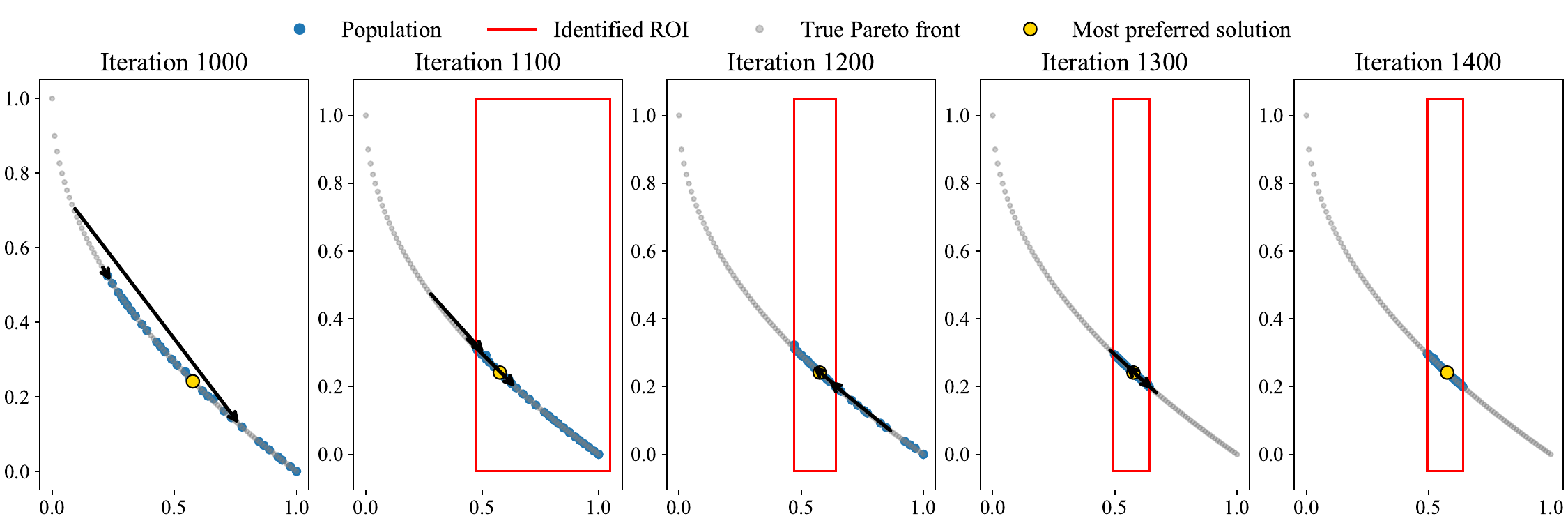}
    \caption{DT-PBO-NSGA's population after each optimization stage. The black arrows indicate which pairwise comparison is asked. Specifically the point at the end of the arrow is preferred over the point at the start of the arrow.}
    \label{fig:process_plot}
\end{figure}

\paragraph{Example}
In Figure~\ref{fig:process_plot}, we show DT-PBO-NSGA's population after each optimization round. The first 1000 generations consist of the initial optimization stage in which the Pareto front is approximated. Afterwards, two randomly sampled pairwise comparisons are presented to the DM and the resulting preference statements are used to train the preference model. The learned preference model partitions the space and influences NSGA's survival operator. After 100 more generations the population has converged to the identified region of interest (i.e. the best leaf) and a new preference elicitation stage starts. In this stage the preference model's active learning module determines the best questions to ask. This process continues until four preference elicitation stages have occurred. Figure~\ref{fig:process_plot} shows the region of interest iteratively reducing in size reaching the most preferred solution.

\section{Experimental Evaluation}\label{sec:numerical_experiments}

\subsection{Experimental Settings}
We perform an extensive empirical study on a large set of benchmark problems to measure the performance of DT-PBO-NSGA as compared to other peer algorithms. We employ artificial DMs that generate the preference statements used by the algorithms. For robustness, we sample 50 different artificial DMs and use them to run 50 repeated trials for each problem. Code to reproduce all experiments is available at: \url{https://github.com/Thomasq99/DT-PBO-NSGA}.

\paragraph*{Peer Algorithms}
We compare our method with two peer methods, one interpretable alternative and one state-of-the-art black-box alternative. First, DT-EMOA~\cite{Shavarani2023}, which is conceptually the closest method to our proposed algorithm. It uses an interpretable decision-tree-based preference model. Specifically, it learns a decision tree over the difference in objective vectors and is therefore less interpretable than our method, which learns a global decision tree directly on the objective space. DT-EMOA uses NSGA-II as the base evolutionary algorithm. Note that there is an extension of DT-EMOA, called RF-EMOA~\cite{Shavarani2025} to include active learning by using a random forest model instead of a decision tree. However, by using an ensemble of models instead of a single decision tree the interpretability is significantly reduced. Therefore, to ensure a comparison with a competitive method that retains a high degree of interpretability, we compare our approach with DT-EMOA rather than RF-EMOA. Finally, DT-EMOA requires a DM to rank a set of $q$ solutions from which $\binom{q}{2}$ pairwise comparisons are generated. To compare to DT-PBO-NSGA and PBO-R-NSGA-III, which both require a DM to answer a set of $j$ pairwise comparisons, we set the number of items that the DM has to rank to $q = j +1$, as under transitivity, a ranking of $n$ items can be represented by $q-1$ pairwise comparisons. 

The second method, PBO-R-NSGA-III~\cite{Taylor2021}, uses a Gaussian process which is learned directly from the pairwise comparisons as preference model. Since Gaussian processes are inherently black-box algorithms, the learned preference model is not interpretable. However, Gaussian processes are data-efficient models that allow for active learning, which makes them very suitable as black-box preference models. Before learning the preference model, PBO-R-NSGA-III uses a balanced reference direction with a large region of interest (determined by R-NSGA-III's $\mu$ parameter, $\mu=1.0$) and runs R-NSGA-III~\cite{vesikar2018reference}. After learning the preference model, the best known point is identified which is then used as the reference direction with a reduced $\mu = 0.1$. 

Note that although all three methods make some assumptions about the preference function (e.g., axis aligned splits for the tree-based algorithms and smoothness assumptions imposed by the covariance kernel for the Gaussian process model), they do not assume explicit functional forms. Methods like NEMO~\cite{Branke2016}, IEMO/D~\cite{Tomczyk2020}, and ERS~\cite{tomczyk2026efficient}, on the other hand, do make these strong assumptions. These methods cannot handle noise or inconsistencies and require the assumed preference model to match the actual preference model. Therefore they are not included as baselines. 

Following Tomczyk and Kadziński~\cite{tomczyk2026efficient}, for each algorithm $G$ initial generations (see Table~\ref{tab:problem_settings}) of the base evolutionary algorithm are executed to get a sufficient approximation of the Pareto front. After that $4$ rounds of preference elicitation happen. The first round happens at generation $G$ and in each round the number of questions asked is dependent on the problem's optimization objectives ($Q_N = M)$. After each preference elicitation round $100$ additional optimization generations are executed to move the population towards the identified region of interest. 

\paragraph*{Algorithm Settings}
To ensure comparability across preference-based evolutionary algorithms, the base evolutionary algorithm NSGA-II and NSGA-III's settings are identical for each compared method.

\begin{enumerate}
    \item The population size was made dependent on the number of objectives $N = 32(M-1)$.
    \item The crossover operator is Simulated Binary Crossover with probability $p_c = 1.0$ and a distribution index $\eta = 30$.
    \item The mutation operator is polynomial mutation with per individual probability $p_m = 0.9$ and distribution index $\eta = 20$. The per variable probability is $1/dv$. For DT-EMOA, the per individual probability is $1.0$, as the implementation does not support other values.
    \item The selection operator is binary tournament selection.
    \item For the first G generations the base evolutionary algorithms use their standard survival operators.
\end{enumerate}

DT-PBO-NSGA's preference model has additional hyperparamers. None of the parameters controlling the tree-growth while training are used. Instead, we rely on DT-PBO's splitting heuristic to control the growth of the tree. For the leaf-fitting the ratio $\sigma_{{noise}}^2 / \sigma_{{prior}}^2$ is set to $\frac{1}{4}$. Finally, for active learning, the hyperparameter $\lambda$ controlling exploration and exploitation is set to a balanced value of $1.5$. An ablation study comparing the performance for different noisy DMs when varying $\lambda$ is included in the supplementary material.

\paragraph*{Benchmark Problems}
To evaluate DT-PBO-NSGA, a diverse set of benchmark problems is employed. The ZDT~\cite{zitzler2000comparison} suite (ZDT5 is excluded due to its discrete decision variables) is used for bi-objective scenarios, providing test cases with convex, non-convex, multimodal, disconnected, and non uniform Pareto fronts. To evaluate performance in higher-dimensional settings, the DTLZ~\cite{deb2002scalable} and WFG~\cite{huband2006review} suites are used. Since holistic pairwise comparisons are only cognitively valid up to 5 dimensions \cite{kivikangas2025effects}, the number of objectives is restricted to $M \in \{3,4,5\}$. The DTLZ suite provides problems with a range of Pareto-front geometries and optimization characteristics, including linear, spherical, degenerate, and disconnected fronts, as well as multimodal landscapes. However, its decision-space constructions are generally relatively regular and often separable. The WFG suite complements DTLZ by introducing more complex decision-space transformations, including bias, deception, non-separability, and parameter dependencies. Together, these benchmark suites enable a comprehensive evaluation of the algorithm's ability to converge toward the Pareto front. Inspired by Tomczyk and Kadziński~\cite{tomczyk2026efficient} the problem settings were set as described in Table~\ref{tab:problem_settings}.

\begin{table}
\caption{Problem settings of each test problem used in the experimental evaluation. $dv$ is the number of decision variables, $k$ is the number of position-related variables and $l$ is the number of distance-related variables. $G$ is the number of generations before the first preference elicitation stage, and $M$ is the number of objectives.}\label{tab:problem_settings}
\centering
\begin{tabular}{@{}lllll@{}}
\toprule
Test Problem & $dv$   & $k$      & $l$  & $G$               \\ \midrule
ZDT1-4       & $30$   &        &    & $1000$            \\
ZDT6         & $10$   &        &    & $1000$            \\
DTLZ1        & $M+4$  &        &    & $500 + 100(M-2)$  \\
DTLZ2        & $M+9$  &        &    & $250 + 50(M-2)$   \\
DTLZ3        & $M+9$  &        &    & $1250 + 250(M-2)$ \\
DTLZ4        & $M+9$  &        &    & $300 + 100(M-2)$  \\
DTLZ5-6      & $M+9$  &        &    & $500 + 100(M-2)$  \\
DTLZ7        & $M+19$ &        &    & $500 + 100(M-2)$  \\
WFG1         & $k+l$  & $2(M-1)$ & $20$ & $1200 + 200(M-2)$ \\
WFG2-9       & $k+l$  & $2(M-1)$ & $20$ & $500 + 100(M-2)$  \\ \bottomrule
\end{tabular}
\end{table}

\paragraph*{Artificial DMs}
Artificial DMs are created from two different kinds of utility functions: Chebyshev and Stewart. The Chebyshev function is defined as $u(\mathbf{x}) = \max_{1 \leq i \leq m} \left\{ w_i \left| f_i(x) - z_i^{ideal} \right| \right\}$, where $\mathbf{z}^* = (z_1^{ideal},..., z_M^{ideal})^T$ is the ideal point and $\mathbf{w} = (w_1, ..,w_M)$ is a randomly sampled weight vector following the Dirichlet distribution. We normalize the objective values via $f_i^{\mathrm{norm}}(x) = \frac{f_i(x)-z_i^{\mathrm{ideal}}}{z_i^{\mathrm{nadir}}-z_i^{\mathrm{ideal}}}$ $\mathbf{f(x)}$ using the ideal and nadir point to ensure all objectives are on similar scales. The Stewart function, as first proposed by Stewart~\cite{stewart1996robustness} and rewritten for minimization problems in Shavarani et al.~\cite{shavarani2023benchmarking} is inspired by psychological studies~\cite{Kahneman1979} and should accurately represent a human decision-maker. There are four different types of Stewart DMs depending on the function’s parameters: 1) a mainly compensatory DM that readily accepts trade-offs between criteria; 2) a mainly compensatory DM that accepts trade-offs but strongly dislikes performance falling below a sharp threshold; 3) a DM with limited compensation for which a broad range of performance below a high reference level is similarly undesirable; and 4) a DM with limited compensation and a sharp preference threshold, making performance below the high reference level particularly undesirable. All four of these types are used in the experiments and the function's parameters are found by randomly sampling a point from the Pareto front, selecting the DM type and solving the optimization problem proposed in Shavarani et al.~\cite{shavarani2023benchmarking} to get the according parameters. Since the Stewart function requires input variables to be between $0$ and $1$ we have to normalize slightly differently. Following Shavarani et al.~\cite{shavarani2023benchmarking}, we normalize values below the nadir point by $0.95 * f_i^{\mathrm{norm}}(x)$. For values above the nadir point we normalize by $0.95 + 0.05 \frac{z_i}{z_i^{worst}}$, where $z_i^{worst}$ is the worst possible value of the $i^{\text{th}}$ objective. Note that for both Chebyshev and Stewart, a solution $\mathbf{x_w}$ attaining a lower utility value $u(\mathbf{x_w})$ is preferred. The algorithms do not have access to these utility functions, instead, they are supplied with preference statements $\mathbf{x_w} \succ \mathbf{x}_l$ generated from the functions.

\subsection{Evaluation}
The performance of each algorithm is evaluated using two metrics: mean utility and best-point utility. Mean utility is defined as the average (mean) utility of the final population, whereas best-point utility is the minimum utility of the final population. Below we focus on results aggregated by dimensionality to provide a concise overview of how performance scales with the number of objectives. Both across runs and across dimensionality, metrics are aggregated by the median instead of the mean to mitigate the effect of outliers caused by bad runs. Additional experimental results are summarized below, with detailed information provided in the Supplementary Material.

\paragraph{Performance across Problem Dimensionality}

To examine how algorithmic performance changes with the problem's number of objectives, Table~\ref{tab:utility_significance} reports the results of the Wilcoxon signed-rank tests grouped by problem dimensionality. Compared with the interpretable baseline DT-EMOA, our method performs particularly well on bi-objective problems, significantly outperforming DT-EMOA in 21 out of 25 cases for average utility and all cases for best-point utility. For three-objective problems, average utility is approximately balanced between the two algorithms, whereas best-point utility strongly favors DT-PBO-NSGA. As the number of objectives increases to four and five, DT-PBO-NSGA is increasingly outperformed in terms of average utility, whilst maintaining a strong advantage in best-point utility. A similar divergence between the two utility measures is observed when comparing DT-PBO-NSGA with PBO-R-NSGA-III, a black-box baseline. For bi-objective problems, neither algorithm shows a clear advantage on either metric. As the number of objectives increases, PBO-R-NSGA-III increasingly outperforms DT-PBO-NSGA in terms of average utility. In contrast, best-point utility increasingly favors DT-PBO-NSGA from three objectives onward. Overall, these results indicate that, as the number of objectives increases, the relative performance of DT-PBO-NSGA in terms of average utility decreases, while its ability to identify high-utility individual solutions remains strong.

\begin{figure}
    \centering
    \includegraphics[width=0.8\linewidth]{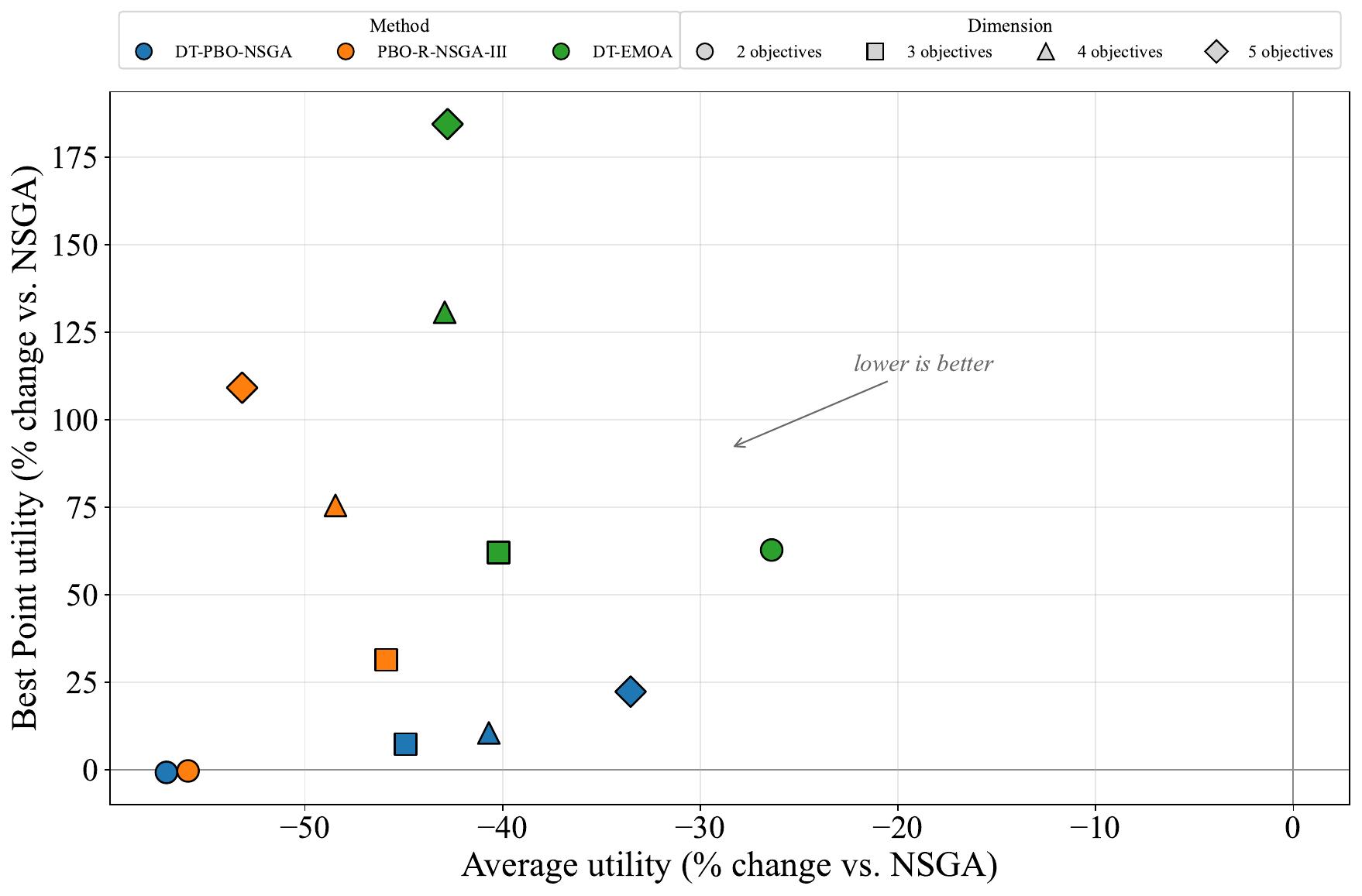}
    \caption{Scatter plot of each method's average and best-point utility relative to NSGA. Results are aggregated across runs and dimensions by taking the median.}
    \label{fig:scatter_utility}
\end{figure}

Figure~\ref{fig:scatter_utility} illustrates this trade-off in greater detail by visualizing the relative performance compared to standard NSGA. Comparing DT-PBO-NSGA to the baselines, we see that, as the dimensionality increases, DT-PBO-NSGA consistently achieves lower best-point utility at the cost of a slightly higher average utility. On average, for bi-objective problems, DT-PBO-NSGA and PBO-R-NSGA-III have similar performance whereas both outperform DT-EMOA. For three-objective problems they both outperform DT-EMOA but DT-PBO-NSGA has a slightly better best point utility and slightly worse average utility than PBO-R-NSGA-III. The trade-off becomes more pronounced in higher dimensions. Specifically, for 4-objective problems, DT-PBO-NSGA reduces the average utility with approximately 41\% at the cost of an increase of best-point utility of 7\%. PBO-R-NSGA-III and DT-EMOA reduce the average utility with 48\% and 43\% respectively, but do so at the cost of 76\% and 131\% worse best point utility. The additional 7 to 2 percentage point reduction in average utility thus comes at the cost of a large increase of best point utility of 69 and 124 percentage points respectively. For 5-objective problems the effect is similar. DT-PBO-NSGA reduces the average utility with 34\% at the cost of a 22\% higher best point utility whilst PBO-R-NSGA-III and DT-EMOA reduce the average utility with 53\% and 43\% at the cost of 109\% and 184\% best point utility respectively. These results indicate that, in higher-dimensional problems, DT-PBO-NSGA offers a somewhat smaller reduction in average utility  but has a significantly higher best-point utility than PBO-R-NSGA-III and DT-EMOA. 

\begin{figure}
    \centering
    \includegraphics[width=0.8\linewidth]{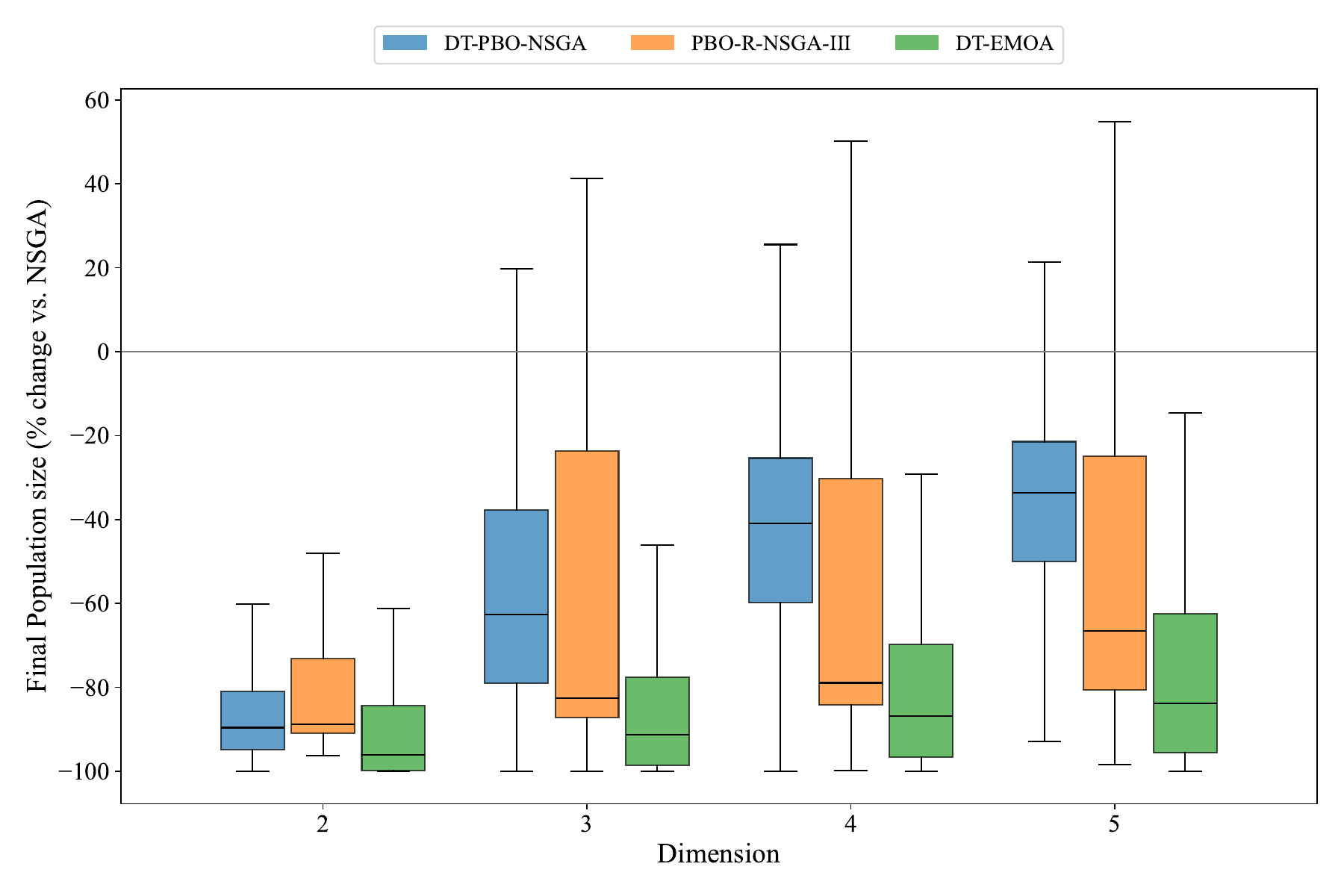}
    \caption{Box plot of the Diagonal of the bounding box spanning the final population. A higher value indicates a larger final region of interest and vice versa.}
    \label{fig:diag_bb}
\end{figure}

These results are caused by DT-PBO-NSGA being more conservative in reducing the size of the final population for higher-dimensional problems. Figure~\ref{fig:diag_bb} shows the reduction of the final population size as compared to standard NSGA. The final population size is measured by the diagonal length of the bounding box spanning the final population for each dimensionality. Compared to the baselines, as the problem's number of objectives increases, DT-PBO-NSGA retains an increasingly larger bounding-box diagonal, indicating that its final population spans a larger region of the objective space. Thus, the higher average utility of DT-PBO-NSGA is caused by preserving a broader set of candidate solutions, while its lower best-point utility indicates that this broader population more often contains the most preferred solution.

Note that all preference-based methods have worse best-point utility than standard NSGA. NSGA consistently identifies at least one solution close to the most preferred solution. In contrast, the preference-based methods restrict their search to an estimated ROI. Even when this ROI closely approximates the true ROI, a small discrepancy between the estimated ROI and the most preferred solution can prevent the final population from containing a solution as preferred as the best solution identified by NSGA.

\begin{table}[t]
\centering
\caption{Pairwise significance testing of DT-PBO-NSGA against competing methods
for average utility and best-point utility. W/T/L denotes the number of cases
where DT-PBO-NSGA significantly outperforms the competitor, shows no significant
difference, or is significantly outperformed, respectively.}
\label{tab:utility_significance}
\begin{tabular}{lccc}
\toprule
Competitor & $D$ & Average utility & Best-Point utility \\
& & (W/T/L) & (W/T/L) \\
\midrule
\multirow{4}{*}{DT-EMOA}
& 2 & 21/4/0  & 25/0/0 \\
& 3 & 16/46/18 & 71/8/1 \\
& 4 & 13/32/35 & 70/9/1 \\
& 5 & 14/25/41 & 72/6/2 \\
\midrule
\multirow{4}{*}{PBO-R-NSGA-III}
& 2 & 7/14/4  & 5/20/0 \\
& 3 & 24/24/32 & 42/28/10 \\
& 4 & 15/21/44 & 54/18/8 \\
& 5 & 9/16/55  & 52/18/10 \\
\bottomrule
\end{tabular}
\end{table}

\paragraph{Performance across Utility Functions}
% To check whether there are performance differences between artificial DMs generated from Stewart utility functions or the Chebyshev utility function we aggregate the results per utility function. In Table~\ref{tab:utility_ranks} we report the average ranks of the different methods based on the average utility and best-point utility. These ranks go from 1-3, as there are three preference-based methods compared. One can see that DT-PBO-NSGA has the lowest best-point utility rank for all utility functions. DT-EMOA performs particularly poorly for Stewart cases 3 and 4, which are characterized by limited compensation and a high reference level. Although decision trees are in principle well suited to capturing such threshold-based preferences, DT-EMOA learns from pairwise differences between objective vectors, thereby losing information about the absolute position of solutions relative to the reference level. Consequently, the learned model cannot adequately represent the threshold- and reference-dependent preferences underlying these utility functions. DT-PBO-NSGA, on the other hand, is trained on the absolute values of the objective vectors and therefore performs well on these functions. 

To check whether performance differs between artificial DMs based on the Stewart and Chebyshev utility functions, we aggregate the results per utility function. Table~\ref{tab:utility_ranks} reports the average ranks of the methods based on the average utility and best-point utility. There is relatively little performance difference between the Chebyshev DM and the Stewart cases 1 and 2 while the limited-compensation cases 3 and 4 differ from the other 3. This is most visible for DT-EMOA, which has its worst best-point ranks on these cases (2.92 and 2.89) and whose average rank drops from best on Chebyshev and Stewart case 2 to worst on Stewart case 3 (2.38) and near worst for Stewart case 4.
Figure~\ref{fig:uf_vis} in the supplementary material helps explain these differences by visualizing each utility function. Stewart cases 3 and 4 have a broad, flat region of near-optimal utility in the middle of the Pareto front and steep increases towards its extremes, where one objective falls below the reference level. The effect of a given change in objective values thus strongly depends on where a solution lies in the objective space. DT-EMOA learns from pairwise differences between objective vectors and therefore cannot represent this positional dependence. For the Chebyshev function and Stewart cases 1 and 2, the utility is closer to locally linear along the front, so pairwise differences remain informative, consistent with DT-EMOA’s good average ranks on these functions. DT-PBO-NSGA is trained on absolute objective values and does not suffer from this limitation, which explains its strong best-point ranks on cases 3 and 4.

\begin{table*}
\centering
\caption{Average ranks of the methods based on average utility and best-point utility across utility functions.}
\label{tab:utility_ranks}
\begin{tabular}{lcc cc cc}
\toprule
& \multicolumn{2}{c}{DT-PBO-NSGA}
& \multicolumn{2}{c}{PBO-R-NSGA-III}
& \multicolumn{2}{c}{DT-EMOA} \\
\cmidrule(lr){2-3}
\cmidrule(lr){4-5}
\cmidrule(lr){6-7}
Utility function
& Avg. rank & Best-point rank
& Avg. rank & Best-point rank
& Avg. rank & Best-point rank \\
\midrule
Chebyshev      & 2.17 & \textbf{1.43} & 1.96 & 2.09 & \textbf{1.87} & 2.47 \\
Stewart case 1 & 2.32 & \textbf{1.32} & \textbf{1.81} & 2.06 & 1.87 & 2.62 \\
Stewart case 2 & 2.42 & \textbf{1.45} & 1.91 & 1.94 & \textbf{1.68} & 2.60 \\
Stewart case 3 & 2.26 & \textbf{1.17} & \textbf{1.36} & 1.91 & 2.38 & 2.92 \\
Stewart case 4 & 2.28 & \textbf{1.21} & \textbf{1.51} & 1.91 & 2.21 & 2.89 \\
\bottomrule
\end{tabular}
\end{table*}

\paragraph{Supplementary Results}
In the supplementary material additional results are presented. First, for each method we test its robustness against noise. In particular, we generate noisy artificial DMs by corrupting their answers with randomly generated Gaussian noise. Results show that all three methods are relatively robust against noise. Second, across various degrees of noise, we investigate the effect of varying DT-PBO-NSGA's $\lambda$ which controls the degree of exploration vs exploitation. As expected, results show that one should decrease $\lambda$, focusing more on exploration, as the level of noise increases. Third, detailed benchmark results showing performance for each individual benchmark problem are presented. These results show largely the same behavior as described in the main experiments. One noteworthy exception is DT-PBO-NSGA's poor performance on DTLZ5 and DTLZ6. For these problems, the Pareto front is effectively reduced to a single curve, for which the curvature is difficult to capture using axis-aligned splits. 
% Fourth, we investigate whether performance differences arise when using different kinds of utility functions. Results show that DT-EMOA performs particularly poor on artificial DMs generated from Stewart type 3 and type 4 utility functions. This is caused by these utility functions being characterized by limited compensation and a high reference level (i.e. being mostly threshold-dependent). Although decision-trees are particularly well-suited for these kinds of functions, DT-EMOA represents pairwise comparisons as the difference in objective vectors thereby losing information on the absolute position of solutions which renders it unable to represent thresholds. DT-PBO-NSGA, on the other hand, is trained on the absolute values of the objective vectors therefore not having this problem and performing well on these kinds of functions. 
Finally, we provide a detailed analysis of each method's performance on the 3- and 5-dimensional DTLZ1, 2, and 7 problems, showing both convergence plots and population plots.

\section{Case Study}\label{sec:case_study}
To demonstrate the interpretability of the learned preference model, we conduct a case study on the real-world RE34 problem~\cite{Tanabe2020}. This engineering problem concerns the crashworthiness of a vehicle and has three easily interpretable objectives, all of which are minimized: vehicle weight, crash-induced acceleration/deceleration, and passenger-compartment intrusion. The lower crash-induced acceleration/deceleration reduces the forces experienced by the occupants, while the lower intrusion into the passenger-compartment reduces the deformation of the occupant space and, consequently, the risk of injury. The weight of the vehicle should be minimized as the goal is a lightweight design.

\begin{figure}
    \centering
    \includegraphics[width=0.8\linewidth]{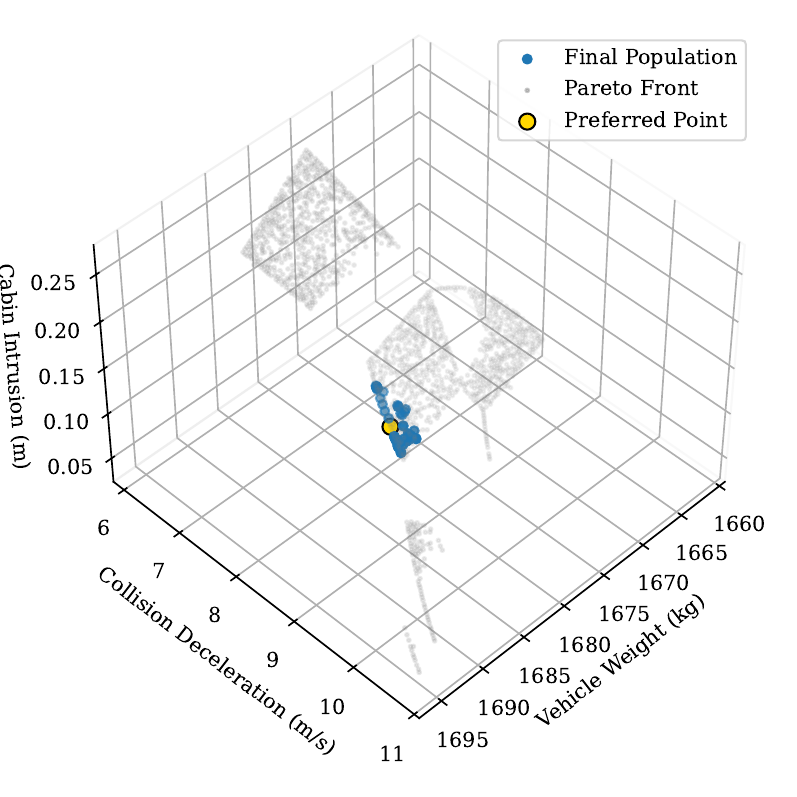}
    \caption{3D scatter plot of the final population identified by DT-PBO-NSGA, the true Pareto front and the decision-maker's most preferred solution for the vehicle crashworthiness design problem.}
    \label{fig:case_study_scatter}
\end{figure}

\begin{figure*}
    \centering
    \includegraphics[width=\linewidth]{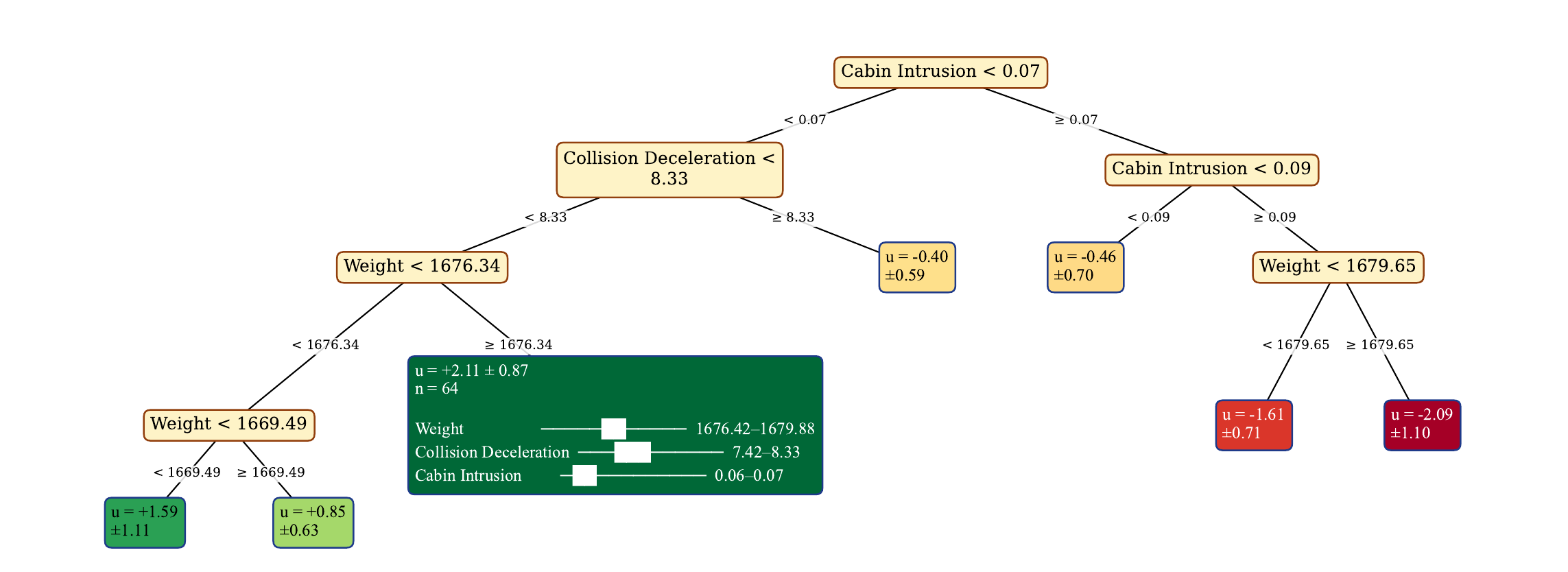}
    \caption{DT-PBO-NSGA's learned preference model. The decision rules show the partitioning created by the tree and each leaf's utility prediction (higher is better) is represented by a Gaussian with a mean and standard deviation. The best leaf contains more information: namely the number of individuals of the final population falling in that leaf and the region spanned by that population, including bar charts representing its size and location w.r.t. the Pareto front.}
    \label{fig:dt_plot}
\end{figure*}

To test our algorithm, we use the same algorithm settings described in the main experiments (thus $4 \times 3$ pairwise comparisons asked) and use a Chebyshev artificial DM. After asking 12 questions in total we get a final population and a learned decision tree. The algorithm's final population can be found in Figure \ref{fig:case_study_scatter}. It shows that the learned ROI is centered at the DM's most preferred point and that it is relatively small compared to the global Pareto front. The learned tree can be found in Figure \ref{fig:dt_plot}. It shows that according to the DM's answers the best vehicles have a cabin intrusion lower than 0.07m, a collision deceleration lower than 8.33 m/s and a vehicle weight higher or equal to 1676.34 kgs. The reason that a higher vehicle weight is preferred as the last decision rule is that allowing for a higher vehicle weight results in a lower Collision Deceleration and/or Cabin Intrusion. The least preferred vehicle designs fall on the right side of the tree and are vehicles with a Cabin intrusion higher or equal to 0.07m. Besides the tree partitioning, Pareto optimality also induces some geometry on the final ROI. This is shown in the best leaf where the final identified ROI is the region with Vehicle weight between 1676.42-1679.88 kgs, Collision deceleration between 7.42-8.33 m/s and Compartment intrusion between 0.06-0.07 meters. Additionally, the bar graphs in the best leaf show the ROI size and location compared to the global Pareto front. For this specific artificial DM the identified ROI focuses mostly on lower Cabin intrusion, medium vehicle weight and low to medium Collision Deceleration. The preference model's interpretability enables the DM to get valuable insights and make changes if desired.

\section{Conclusion}\label{sec:conclusion}

We proposed a novel interactive evolutionary algorithm DT-PBO-NSGA that uses an interpretable tree-based preference model. The model is learned from pairwise comparisons and supports active learning. Its decision-tree structure provides an interpretable partitioning of the objective space based on the DM's preferences. Each leaf is ranked and this information is used to adapt NSGA's survival and mating selection operators to guide the population towards the identified ROI. Specifically, a preference-based sorting step is added in between NSGA's dominance-based and diversity-based sorting.

Extensive experiments across the ZDT, DTLZ, and WFG benchmark suites comparing DT-PBO-NSGA with another interpretable alternative DT-EMOA and a black-box algorithm PBO-R-NSGA-III showed that DT-PBO-NSGA performs particularly well on lower-dimensional problems and maintains a favorable trade-off as the number of objectives increases. While its performance in terms of average utility is slightly worse than DT-EMOA and PBO-R-NSGA-III in higher dimensions, DT-PBO-NSGA has a much better best-point utility. This behavior is mainly caused by its more conservative reduction of the final population, preserving a broader region of the objective space and thereby increasing the likelihood of finding the best solution. 
Furthermore, results aggregated per utility function show that DT-EMOA performs particularly poorly on Stewart utility functions of type 3 and 4. DT-EMOA cannot model these based on the difference in objective vectors. DT-PBO-NSGA, on the other hand, learns directly from the objective vectors and can therefore model these positional dependencies. 
Overall, these results show that DT-PBO-NSGA is competitive and interpretable, particularly when finding the most preferred solution is more important than finding a set of final solutions that are all good.
Additionally, a case study on the real-world RE34 vehicle crashworthiness problem demonstrated the interpretability of the proposed approach. DT-PBO-NSGA identified an ROI close to the most preferred solution, while the learned decision tree revealed that the artificial DM's preferences were primarily driven by minimizing cabin intrusion.

Future work could focus on many interesting directions. First, future work could extend decomposition- and indicator-based algorithms with DT-PBO's preference model. Secondly, the tree-based preference model makes axis-aligned splits and can thus have difficulty accommodating preference functions that do not follow these axes. Third, the learned decision tree does not inherently enforce monotonicity, even though preferences in MOPs are generally assumed to be monotonic with respect to the objectives. Incorporating this assumption directly into the model could improve learning efficiency by eliminating the need to infer these relationships from the preference data. Finally, experiments using real-world decision-makers can be done to validate the learned preference model's interpretability and its performance in real-life settings.

% Wang, H., Branke, J. and Poloczek, M. (2025) 'Bayesian optimization with preference exploration using a monotonic neural network ensemble', in The Thirty-ninth Annual Conference on Neural Information Processing Systems.

% In prior work, Lin et al. [2022] assume that during the Preference Exploration stage, experimenters
% could present DMs with arbitrary output pairs, including those that were unobserved or even nonexistent. However, presenting such unattainable solutions risks causing disappointment when DMs
% later discover these options are unavailable. Therefore, our work constrains experimenters to only
% present output pairs that have been previously observed.
\newpage

\bibliographystyle{IEEEtran}
\bibliography{final_bibliography}

% Biography text here without a photo.

% \begin{IEEEbiography}[{\includegraphics[width=1in,height=1.25in,clip,keepaspectratio]{fig1.png}}]{IEEE Publications Technology Team}
% In this paragraph you can place your educational, professional background and research and other interests.\end{IEEEbiography}

\appendices

\section{Additional Experiments}
In this appendix, we provide experiments testing the algorithm's robustness against noise, provide detailed test results on the benchmark suites ZDT, WFG, DTLZ by only aggregating on the utility function, provide test results per utility function, provide an ablation study of $\lambda$, including performance when varying the DM's noise, and finally provide a detailed analysis of the results on DTLZ1,2, and 7, including convergence and population plots.

\subsection{Robustness against Noise}
\begin{figure}
    \centering
    \includegraphics[width=\linewidth]{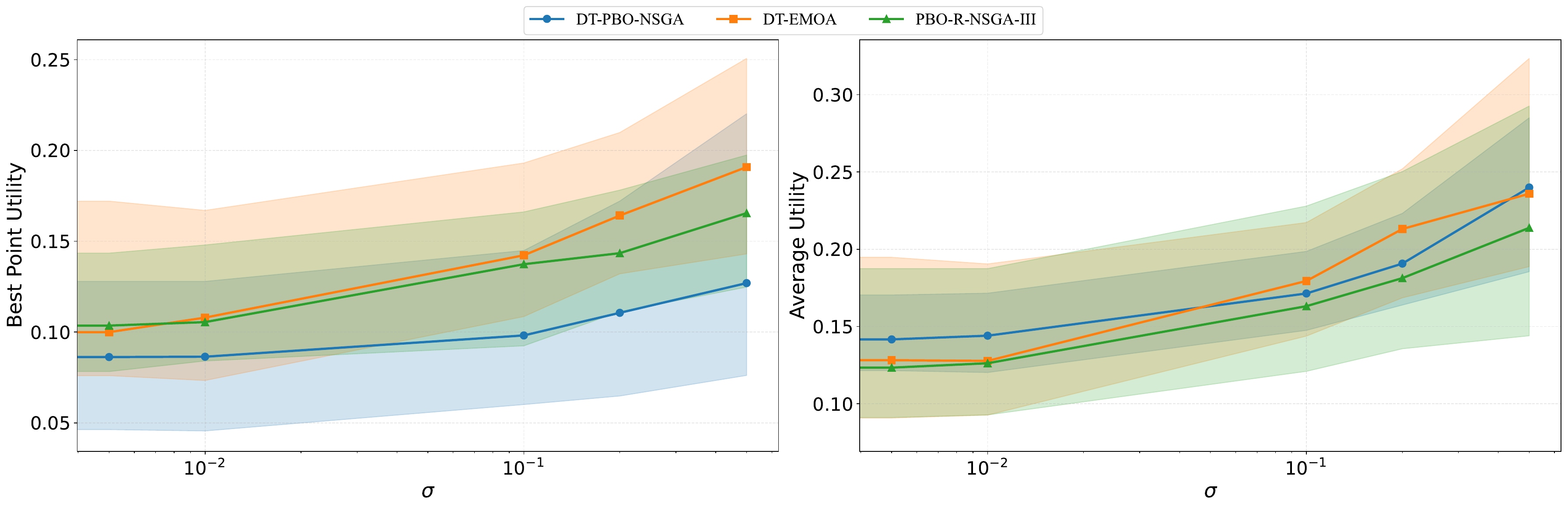}
    \caption{Line plots showing each method's robustness against noise. Results average all problems and utility functions. The error bars denote one standard deviation.}
    \label{fig:noise}
\end{figure}

In practice, there is no guarantee that DMs give consistent answers. Therefore, we test each algorithm's robustness against inconsistencies. To do so, we introduce Gaussian noise into the artificial DM's value function, such that $(v_{\text{noise}} = v + \epsilon)$, where $\epsilon \sim \mathcal{N}(0,\sigma^2)$. For each object, noise is independently sampled, and the resulting noisy values are used to determine the DM's preference statement. We repeat the experiments for different values of \(\sigma\), with larger values corresponding to increasingly noisy DM responses. The results are shown in Figure~\ref{fig:noise}. Although performance generally decreases as the noise level increases, all three methods, including DT-PBO-NSGA, remain relatively robust to moderate levels of noise.

\subsection{Lambda ablation study}
\begin{figure}
    \centering
    \includegraphics[width=\linewidth]{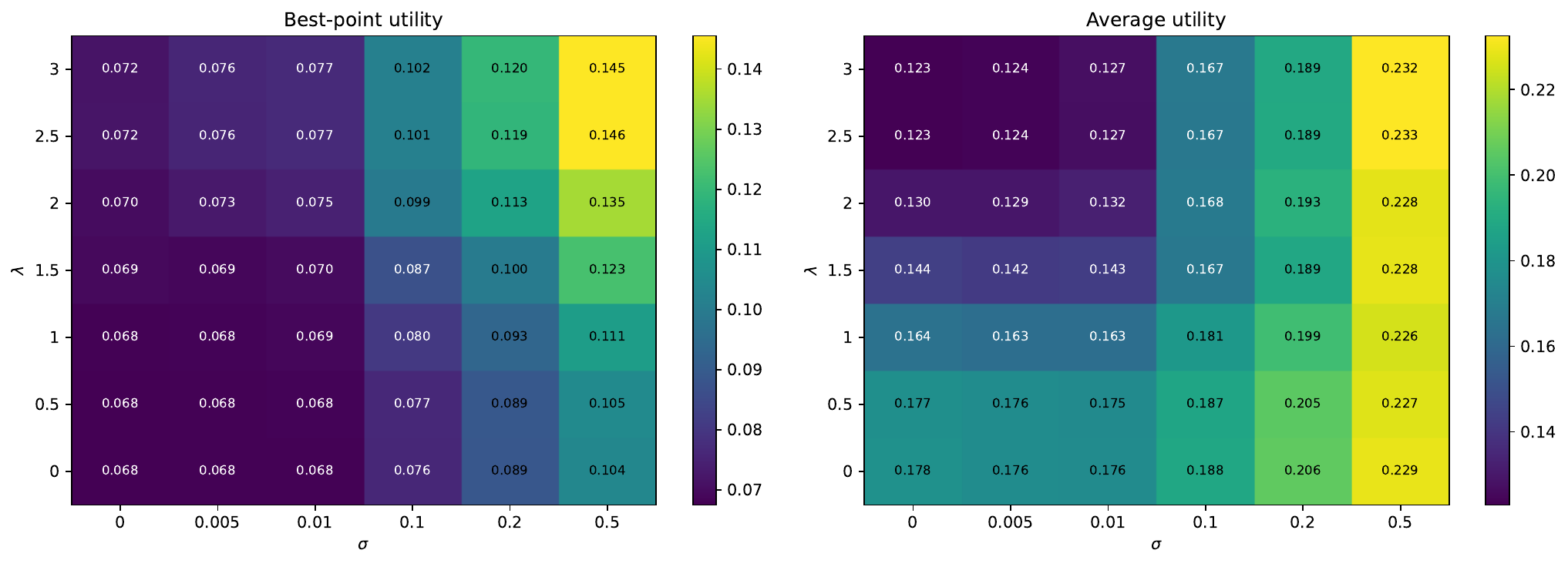}
    \caption{Heat maps showing DT-PBO-NSGA's performance in both best-point utility and average utility when varying both $\lambda$ and noise $\sigma$.}
    \label{fig:lambda_ablation}
\end{figure}

DT-PBO-NSGA has a hyperparameter $\lambda$ that determines the level of exploration vs exploitation. A lower value of $\lambda$ focuses more on exploration and a higher value on exploitation. Figure~\ref{fig:lambda_ablation} shows the averaged (by median) best-point utility and average utility (lower is better) of all problems and utility functions when varying both the level of noise $\sigma$ and $\lambda$. For one noise level, one can see that the average utility decreases when $\lambda$ decreases. For the best-point utility this is reversed. This is expected, as with a higher $\lambda$ more exploitation happens, which reduces the size of the region of interest which in turn reduces the average utility. At the same time, reducing the size of the region of interest can cause it to miss the most preferred solution, therefore, increasing the best-point utility. For noiseless DMs, a very high $\lambda$ has the best tradeoff, compared to a balanced $\lambda = 1.5$ the median best-point utility is worsened by $0.03$ but the average utility is improved by $0.21$. For noisy decision-makers (e.g. $\sigma=0.1$), a very high $\lambda$ is generally outperformed by a conservative or balanced $\lambda$. For example, $\lambda = 1.5$ improves the best-point utility by $0.15$ whilst having the same average utility. The optimal choice of $\lambda$ thus depends on the level of inconsistencies present in the DM's answers. 

\subsection{Synthetic Benchmark Results Aggregated per Problem}
For completeness, we have included detailed benchmark results. Specifically, for each benchmark problem we have aggregated on the utility functions and runs. The medians of the average utility of the final population, standard deviations of average utility, and average ranks are shown in Table~\ref{tab:average-utility}. Similarly, for the best-point utility the results can be found in Table~\ref{tab:best-point-utility}. The results largely follow the same trend as presented in the main paper. Namely, compared to DT-PBO-NSGA, as the problem dimensionality increases, both PBO-R-NSGA-III and DT-EMOA get an increasingly better average utility but also an increasingly worse best-point utility. The main exception to the trend are the problems DTLZ5 and DTLZ6, for which DT-PBO-NSGA performs poorly. These problems both have a Pareto front geometry that largely reduces to a curve in high-dimensional space. Since DT-PBO-NSGA makes axis-aligned splits it has trouble making rectangular splits that follow the curvature of the Pareto front.

\begin{table*}
\centering
\caption{Average utility comparison on ZDT, DTLZ, and WFG benchmark problems. Lower median utility is better. For each row the best utility and rank are made bold.}
\label{tab:average-utility}

\scriptsize
\setlength{\tabcolsep}{2.5pt}
\renewcommand{\arraystretch}{0.85}

\begin{tabular}{ccccc|ccc|ccc}
\toprule
\multirow{2}{*}{Problem} &
\multirow{2}{*}{$M$} &
\multicolumn{3}{c}{DT-PBO-NSGA} &
\multicolumn{3}{c}{PBO-R-NSGA-III} &
\multicolumn{3}{c}{DT-EMOA} \\
\cmidrule(lr){3-5}
\cmidrule(lr){6-8}
\cmidrule(lr){9-11}
& &
Median $\downarrow$ & Std & Rank $\downarrow$ &
Median $\downarrow$ & Std & Rank $\downarrow$ &
Median $\downarrow$ & Std & Rank $\downarrow$ \\
\midrule

% ==================== ZDT ====================
ZDT1 & 2 &
0.1396 & 0.1693 & 1.6760 &
\textbf{0.1379} & 0.1648 & \textbf{1.5800} &
0.3178 & 0.2330 & 2.7440 \\

ZDT2 & 2 &
0.2450 & 0.1799 & \textbf{1.8120} &
\textbf{0.2398} & 0.1633 & 1.8600 &
0.3408 & 0.1923 & 2.3280 \\

ZDT3 & 2 &
\textbf{0.1657} & 0.1946 & \textbf{1.5280} &
0.1973 & 0.1723 & 2.1560 &
0.2465 & 0.2036 & 2.3160 \\

ZDT4 & 2 &
\textbf{0.1405} & 0.1733 & 1.7520 &
0.1460 & 0.1693 & \textbf{1.4840} &
0.3067 & 0.2365 & 2.7640 \\

ZDT6 & 2 &
\textbf{0.2120} & 0.1835 & \textbf{1.4920} &
0.2748 & 0.1936 & 1.9560 &
0.3019 & 0.1897 & 2.5520 \\

\midrule

% ==================== DTLZ ====================
\multirow{3}{*}{DTLZ1}
& 3 & 0.1204 & 0.1225 & 2.3360
& \textbf{0.0732} & 0.1106 & \textbf{1.1720}
& 0.2417 & 0.1499 & 2.4920 \\
& 4 & 0.1019 & 0.1052 & 2.3120
& \textbf{0.0365} & 0.0734 & \textbf{1.1840}
& 0.1693 & 0.1149 & 2.5040 \\
& 5 & 0.0798 & 0.0881 & 2.2040
& \textbf{0.0302} & 0.0523 & \textbf{1.1240}
& 0.1666 & 0.5026 & 2.6720 \\

\multirow{3}{*}{DTLZ2}
& 3 & 0.1886 & 0.1611 & 2.4880
& \textbf{0.1081} & 0.1144 & \textbf{1.3360}
& 0.1885 & 0.1557 & 2.1760 \\
& 4 & 0.1624 & 0.1332 & 2.6320
& \textbf{0.0509} & 0.0754 & \textbf{1.2120}
& 0.1416 & 0.1274 & 2.1560 \\
& 5 & 0.1309 & 0.1230 & 2.4680
& \textbf{0.0444} & 0.0617 & \textbf{1.2840}
& 0.1355 & 0.1091 & 2.2480 \\

\multirow{3}{*}{DTLZ3}
& 3 & 0.1925 & 0.1671 & 2.4360
& \textbf{0.1134} & 0.1083 & \textbf{1.3880}
& 0.2218 & 0.1605 & 2.1760 \\
& 4 & 0.1566 & 0.1279 & 2.3480
& \textbf{0.0605} & 0.0738 & \textbf{1.1400}
& 0.2169 & 0.1981 & 2.5120 \\
& 5 & 0.1303 & 0.1167 & 2.3320
& \textbf{0.0490} & 0.0565 & \textbf{1.1600}
& 0.2015 & 0.8844 & 2.5080 \\

\multirow{3}{*}{DTLZ4}
& 3 & 0.2230 & 0.1493 & 2.2531
& \textbf{0.1636} & 0.1270 & 1.9918
& 0.1656 & 0.1636 & \textbf{1.7551} \\
& 4 & 0.1476 & 0.1408 & 2.4816
& \textbf{0.0715} & 0.0706 & \textbf{1.6367}
& 0.1116 & 0.1110 & 1.8816 \\
& 5 & 0.1287 & 0.1145 & 2.5640
& \textbf{0.0600} & 0.0578 & \textbf{1.5800}
& 0.0959 & 0.0947 & 1.8560 \\

\multirow{3}{*}{DTLZ5}
& 3 & \textbf{0.3269} & 0.1626 & \textbf{1.6560}
& 0.3343 & 0.1480 & 2.3840
& 0.3428 & 0.1873 & 1.9600 \\
& 4 & 0.3401 & 0.1838 & 2.0160
& \textbf{0.3072} & 0.1367 & 2.0480
& 0.3498 & 0.2185 & \textbf{1.9360} \\
& 5 & 0.3969 & 0.1999 & 2.3040
& 0.3600 & 0.1289 & 1.8880
& \textbf{0.2966} & 0.2376 & \textbf{1.8080} \\

\multirow{3}{*}{DTLZ6}
& 3 & \textbf{0.3597} & 0.1756 & \textbf{1.8940}
& 0.3801 & 0.1825 & 2.1400
& 0.3946 & 0.1955 & 1.9660 \\
& 4 & 0.4731 & 0.3182 & 2.2160
& \textbf{0.3464} & 0.1894 & \textbf{1.6760}
& 0.4804 & 0.3433 & 2.1080 \\
& 5 & 0.4999 & 0.1933 & 2.4380
& \textbf{0.3412} & 0.1491 & \textbf{1.4840}
& 0.4611 & 0.4926 & 2.0780 \\

\multirow{3}{*}{DTLZ7}
& 3 & 0.2906 & 0.1706 & 2.4880
& \textbf{0.1972} & 0.1599 & \textbf{1.6280}
& 0.2271 & 0.1964 & 1.8840 \\
& 4 & 0.2196 & 0.1853 & 2.6160
& \textbf{0.1463} & 0.1744 & \textbf{1.4920}
& 0.1890 & 0.1774 & 1.8920 \\
& 5 & 0.2142 & 0.1565 & 2.6880
& \textbf{0.1294} & 0.1452 & \textbf{1.3800}
& 0.1933 & 0.1600 & 1.9320 \\

\midrule

% ==================== WFG ====================
\multirow{3}{*}{WFG1}
& 3 & \textbf{0.1393} & 0.2029 & \textbf{1.7160}
& 0.2327 & 0.2016 & 2.3960
& 0.2082 & 0.1878 & 1.8880 \\
& 4 & \textbf{0.1149} & 0.1901 & 1.7920
& 0.2091 & 0.1697 & 2.5240
& 0.1390 & 0.1619 & \textbf{1.6840} \\
& 5 & \textbf{0.0878} & 0.1632 & 1.7120
& 0.1707 & 0.1545 & 2.6120
& 0.1143 & 0.1435 & \textbf{1.6760} \\

\multirow{3}{*}{WFG2}
& 3 & \textbf{0.0939} & 0.1234 & \textbf{1.8800}
& 0.1055 & 0.1412 & 1.9760
& 0.1147 & 0.1322 & 2.1440 \\
& 4 & 0.0634 & 0.0910 & \textbf{1.8600}
& \textbf{0.0571} & 0.1222 & 1.9920
& 0.0762 & 0.1286 & 2.1480 \\
& 5 & 0.0442 & 0.0613 & 1.9160
& \textbf{0.0336} & 0.0772 & \textbf{1.6320}
& 0.0848 & 0.0901 & 2.4520 \\

\multirow{3}{*}{WFG3}
& 3 & \textbf{0.2495} & 0.1780 & \textbf{1.6600}
& 0.3022 & 0.1799 & 2.1480
& 0.3947 & 0.1949 & 2.1920 \\
& 4 & \textbf{0.2925} & 0.1737 & \textbf{1.5840}
& 0.4480 & 0.1546 & 2.4200
& 0.4147 & 0.2072 & 1.9960 \\
& 5 & 0.4493 & 0.1824 & 1.9320
& 0.4902 & 0.1871 & 2.3840
& \textbf{0.3935} & 0.2390 & \textbf{1.6840} \\

\multirow{3}{*}{WFG4}
& 3 & 0.2007 & 0.1718 & 2.1720
& 0.2389 & 0.1867 & 2.1120
& \textbf{0.1634} & 0.1571 & \textbf{1.7160} \\
& 4 & 0.1684 & 0.1374 & 2.2880
& 0.1504 & 0.1547 & 2.0320
& \textbf{0.1026} & 0.1257 & \textbf{1.6800} \\
& 5 & 0.1536 & 0.1145 & 2.4440
& 0.1138 & 0.1416 & 1.9880
& \textbf{0.0697} & 0.0877 & \textbf{1.5680} \\

\multirow{3}{*}{WFG5}
& 3 & 0.1877 & 0.1615 & 2.1440
& 0.1941 & 0.2201 & 2.0560
& \textbf{0.1833} & 0.1615 & \textbf{1.8000} \\
& 4 & 0.1509 & 0.1463 & 2.2240
& 0.1391 & 0.2202 & 2.0680
& \textbf{0.1104} & 0.1311 & \textbf{1.7080} \\
& 5 & 0.1286 & 0.1286 & 2.2480
& 0.1048 & 0.1987 & 2.1040
& \textbf{0.0844} & 0.1124 & \textbf{1.6480} \\

\multirow{3}{*}{WFG6}
& 3 & 0.1937 & 0.1538 & 2.2680
& 0.1919 & 0.1754 & 1.8840
& \textbf{0.1811} & 0.1454 & \textbf{1.8480} \\
& 4 & 0.1601 & 0.1396 & 2.3520
& 0.1262 & 0.1728 & 1.8800
& \textbf{0.1089} & 0.1129 & \textbf{1.7680} \\
& 5 & 0.1455 & 0.1319 & 2.5280
& 0.1026 & 0.1497 & 1.7600
& \textbf{0.0926} & 0.0919 & \textbf{1.7120} \\

\multirow{3}{*}{WFG7}
& 3 & 0.1908 & 0.1446 & 2.2000
& \textbf{0.1699} & 0.1743 & \textbf{1.9000}
& 0.1890 & 0.1588 & \textbf{1.9000} \\
& 4 & 0.1519 & 0.1294 & 2.2080
& 0.1307 & 0.1848 & 1.9760
& \textbf{0.1184} & 0.1295 & \textbf{1.8160} \\
& 5 & 0.1367 & 0.1367 & 2.4400
& \textbf{0.0932} & 0.1699 & 1.8680
& 0.0951 & 0.1070 & \textbf{1.6920} \\

\multirow{3}{*}{WFG8}
& 3 & 0.2121 & 0.1588 & 2.0480
& 0.2624 & 0.2262 & 2.2760
& \textbf{0.1949} & 0.1493 & \textbf{1.6760} \\
& 4 & 0.1622 & 0.1456 & 2.3400
& 0.1469 & 0.1950 & 1.9760
& \textbf{0.1227} & 0.1324 & \textbf{1.6840} \\
& 5 & 0.1425 & 0.1462 & 2.4480
& 0.1094 & 0.1577 & 1.8760
& \textbf{0.1034} & 0.1111 & \textbf{1.6760} \\

\multirow{3}{*}{WFG9}
& 3 & 0.2079 & 0.1675 & 2.1840
& 0.2122 & 0.1696 & \textbf{1.9000}
& \textbf{0.2008} & 0.1651 & 1.9160 \\
& 4 & 0.1488 & 0.1570 & 2.3560
& 0.1506 & 0.1307 & 1.8880
& \textbf{0.1289} & 0.1303 & \textbf{1.7560} \\
& 5 & 0.1533 & 0.1319 & 2.5360
& \textbf{0.0921} & 0.1003 & \textbf{1.6400}
& 0.1048 & 0.1047 & 1.8240 \\

\bottomrule
\end{tabular}
\end{table*}

\begin{table*}
\centering
\caption{Best-point utility comparison on ZDT, DTLZ, and WFG benchmark problems. Lower median utility is better. For each row the best utility and rank are made bold.}
\label{tab:best-point-utility}

\scriptsize
\setlength{\tabcolsep}{2.5pt}
\renewcommand{\arraystretch}{0.85}

\begin{tabular}{ccccc|ccc|ccc}
\toprule
\multirow{2}{*}{Problem} &
\multirow{2}{*}{$M$} &
\multicolumn{3}{c}{DT-PBO-NSGA} &
\multicolumn{3}{c}{PBO-R-NSGA-III} &
\multicolumn{3}{c}{DT-EMOA} \\
\cmidrule(lr){3-5}
\cmidrule(lr){6-8}
\cmidrule(lr){9-11}
& &
Median $\downarrow$ & Std & Rank $\downarrow$ &
Median $\downarrow$ & Std & Rank $\downarrow$ &
Median $\downarrow$ & Std & Rank $\downarrow$ \\
\midrule

% ==================== ZDT ====================
ZDT1 & 2 &
0.1228 & 0.1668 & 1.6040 &
\textbf{0.1180} & 0.1642 & \textbf{1.4800} &
0.2673 & 0.2279 & 2.9160 \\

ZDT2 & 2 &
\textbf{0.2274} & 0.1785 & \textbf{1.6840} &
0.2319 & 0.1644 & 1.8400 &
0.3020 & 0.1951 & 2.4760 \\

ZDT3 & 2 &
\textbf{0.1493} & 0.1939 & \textbf{1.6120} &
0.1507 & 0.1801 & 1.6640 &
0.2218 & 0.2021 & 2.7240 \\

ZDT4 & 2 &
0.1318 & 0.1711 & 1.6200 &
\textbf{0.1307} & 0.1689 & \textbf{1.5520} &
0.2640 & 0.2368 & 2.8280 \\

ZDT6 & 2 &
\textbf{0.2044} & 0.1856 & \textbf{1.5360} &
0.2594 & 0.1901 & 1.7320 &
0.2931 & 0.1850 & 2.7320 \\

\midrule

% ==================== DTLZ ====================
\multirow{3}{*}{DTLZ1}
& 3 & \textbf{0.0632} & 0.1106 & \textbf{1.5440}
& 0.0655 & 0.1092 & 1.6400
& 0.2165 & 0.1516 & 2.8160 \\
& 4 & 0.0376 & 0.0739 & \textbf{1.4680}
& \textbf{0.0346} & 0.0712 & 1.7320
& 0.1481 & 0.1175 & 2.8000 \\
& 5 & \textbf{0.0256} & 0.0465 & \textbf{1.3960}
& \textbf{0.0256} & 0.0482 & 1.7080
& 0.1161 & 0.1833 & 2.8960 \\

\multirow{3}{*}{DTLZ2}
& 3 & 0.1052 & 0.1210 & \textbf{1.6920}
& \textbf{0.1003} & 0.1111 & 1.8440
& 0.1478 & 0.1448 & 2.4640 \\
& 4 & 0.0546 & 0.0825 & \textbf{1.6120}
& \textbf{0.0450} & 0.0738 & 1.7760
& 0.0917 & 0.1162 & 2.6120 \\
& 5 & \textbf{0.0286} & 0.0657 & \textbf{1.6080}
& 0.0333 & 0.0540 & 1.7240
& 0.0969 & 0.0981 & 2.6680 \\

\multirow{3}{*}{DTLZ3}
& 3 & 0.1009 & 0.1198 & \textbf{1.6560}
& \textbf{0.1002} & 0.1035 & 1.8160
& 0.2041 & 0.1555 & 2.5280 \\
& 4 & \textbf{0.0449} & 0.0893 & \textbf{1.5000}
& 0.0508 & 0.0730 & 1.6880
& 0.1739 & 0.1599 & 2.8120 \\
& 5 & \textbf{0.0325} & 0.0598 & \textbf{1.4840}
& 0.0329 & 0.0533 & 1.6440
& 0.1549 & 0.3199 & 2.8720 \\

\multirow{3}{*}{DTLZ4}
& 3 & 0.1341 & 0.1417 & \textbf{1.6122}
& 0.1490 & 0.1281 & 2.2408
& \textbf{0.1307} & 0.1417 & 2.1469 \\
& 4 & \textbf{0.0579} & 0.0861 & \textbf{1.6857}
& 0.0605 & 0.0695 & 2.0612
& 0.0672 & 0.0974 & 2.2531 \\
& 5 & \textbf{0.0305} & 0.0626 & \textbf{1.6240}
& 0.0359 & 0.0541 & 1.9520
& 0.0637 & 0.0810 & 2.4240 \\

\multirow{3}{*}{DTLZ5}
& 3 & 0.2939 & 0.1646 & 1.8560
& \textbf{0.2867} & 0.1509 & \textbf{1.8120}
& 0.3086 & 0.1796 & 2.3320 \\
& 4 & 0.2158 & 0.1617 & 1.8520
& \textbf{0.2052} & 0.1321 & \textbf{1.8120}
& 0.3055 & 0.1945 & 2.3360 \\
& 5 & 0.1869 & 0.1538 & 1.9100
& \textbf{0.1803} & 0.1109 & \textbf{1.5520}
& 0.2507 & 0.2099 & 2.5380 \\

\multirow{3}{*}{DTLZ6}
& 3 & 0.3230 & 0.1706 & 2.1880
& \textbf{0.2944} & 0.1515 & \textbf{1.1360}
& 0.3803 & 0.1917 & 2.6760 \\
& 4 & 0.2342 & 0.1658 & 2.1340
& \textbf{0.2061} & 0.1221 & \textbf{1.3400}
& 0.4221 & 0.2823 & 2.5260 \\
& 5 & 0.2311 & 0.1593 & 2.2040
& \textbf{0.1772} & 0.0957 & \textbf{1.1980}
& 0.3495 & 0.3876 & 2.5980 \\

\multirow{3}{*}{DTLZ7}
& 3 & 0.1584 & 0.1764 & 1.7720
& \textbf{0.1557} & 0.1650 & \textbf{1.6200}
& 0.2043 & 0.1927 & 2.6080 \\
& 4 & \textbf{0.1002} & 0.1726 & \textbf{1.7080}
& 0.1034 & 0.1763 & 1.8040
& 0.1567 & 0.1766 & 2.4880 \\
& 5 & 0.0994 & 0.1380 & 1.7440
& \textbf{0.0942} & 0.1436 & \textbf{1.6440}
& 0.1596 & 0.1544 & 2.6120 \\

\midrule

% ==================== WFG ====================
\multirow{3}{*}{WFG1}
& 3 & \textbf{0.1105} & 0.1902 & \textbf{1.1560}
& 0.1565 & 0.1984 & 2.3920
& 0.1986 & 0.1909 & 2.4520 \\
& 4 & \textbf{0.0863} & 0.1663 & \textbf{1.3160}
& 0.1532 & 0.1736 & 2.5880
& 0.1324 & 0.1655 & 2.0960 \\
& 5 & \textbf{0.0727} & 0.1459 & \textbf{1.3600}
& 0.1326 & 0.1567 & 2.8120
& 0.0952 & 0.1449 & 1.8280 \\

\multirow{3}{*}{WFG2}
& 3 & \textbf{0.0525} & 0.0955 & \textbf{1.5560}
& 0.0701 & 0.1156 & 1.9160
& 0.0915 & 0.1224 & 2.5280 \\
& 4 & \textbf{0.0190} & 0.0493 & \textbf{1.3240}
& 0.0372 & 0.0892 & 2.1360
& 0.0602 & 0.1212 & 2.5400 \\
& 5 & \textbf{0.0108} & 0.0304 & \textbf{1.2680}
& 0.0190 & 0.0589 & 2.0920
& 0.0517 & 0.0853 & 2.6400 \\

\multirow{3}{*}{WFG3}
& 3 & 0.1933 & 0.1843 & 1.7680
& \textbf{0.1914} & 0.1905 & \textbf{1.7480}
& 0.3826 & 0.2006 & 2.4840 \\
& 4 & \textbf{0.1564} & 0.1780 & \textbf{1.5720}
& 0.1566 & 0.2062 & 2.2000
& 0.4031 & 0.2066 & 2.2280 \\
& 5 & \textbf{0.1979} & 0.1752 & \textbf{1.5680}
& 0.2447 & 0.1856 & 2.3480
& 0.3758 & 0.2338 & 2.0840 \\

\multirow{3}{*}{WFG4}
& 3 & \textbf{0.0878} & 0.1181 & \textbf{1.5840}
& 0.1999 & 0.1786 & 2.4000
& 0.1122 & 0.1552 & 2.0160 \\
& 4 & \textbf{0.0458} & 0.0682 & \textbf{1.4960}
& 0.1259 & 0.1495 & 2.4800
& 0.0702 & 0.1186 & 2.0240 \\
& 5 & \textbf{0.0268} & 0.0399 & \textbf{1.4200}
& 0.0941 & 0.1336 & 2.5360
& 0.0424 & 0.0821 & 2.0440 \\

\multirow{3}{*}{WFG5}
& 3 & \textbf{0.1019} & 0.1182 & \textbf{1.5320}
& 0.1717 & 0.2044 & 2.3560
& 0.1408 & 0.1464 & 2.1120 \\
& 4 & \textbf{0.0511} & 0.0745 & \textbf{1.4600}
& 0.1173 & 0.2037 & 2.4440
& 0.0715 & 0.1180 & 2.0960 \\
& 5 & \textbf{0.0269} & 0.0537 & \textbf{1.4000}
& 0.0835 & 0.1765 & 2.5200
& 0.0515 & 0.0916 & 2.0800 \\

\multirow{3}{*}{WFG6}
& 3 & \textbf{0.1023} & 0.1181 & \textbf{1.6200}
& 0.1695 & 0.1559 & 2.2240
& 0.1453 & 0.1404 & 2.1560 \\
& 4 & \textbf{0.0471} & 0.0903 & \textbf{1.4760}
& 0.1001 & 0.1537 & 2.2960
& 0.0875 & 0.1029 & 2.2280 \\
& 5 & \textbf{0.0370} & 0.0613 & \textbf{1.4040}
& 0.0829 & 0.1282 & 2.3880
& 0.0691 & 0.0845 & 2.2080 \\

\multirow{3}{*}{WFG7}
& 3 & \textbf{0.1028} & 0.1149 & \textbf{1.6000}
& 0.1429 & 0.1622 & 2.0760
& 0.1606 & 0.1546 & 2.3240 \\
& 4 & \textbf{0.0435} & 0.0830 & \textbf{1.4880}
& 0.1089 & 0.1638 & 2.3200
& 0.0969 & 0.1176 & 2.1920 \\
& 5 & \textbf{0.0334} & 0.0640 & \textbf{1.4040}
& 0.0738 & 0.1477 & 2.4120
& 0.0741 & 0.1046 & 2.1840 \\

\multirow{3}{*}{WFG8}
& 3 & \textbf{0.1237} & 0.1285 & \textbf{1.6960}
& 0.2166 & 0.2003 & 2.3280
& 0.1717 & 0.1451 & 1.9760 \\
& 4 & \textbf{0.0607} & 0.0908 & \textbf{1.6000}
& 0.1159 & 0.1614 & 2.3120
& 0.0996 & 0.1254 & 2.0880 \\
& 5 & \textbf{0.0445} & 0.0713 & \textbf{1.5560}
& 0.0805 & 0.1234 & 2.3560
& 0.0709 & 0.1028 & 2.0880 \\

\multirow{3}{*}{WFG9}
& 3 & \textbf{0.1064} & 0.1334 & \textbf{1.5840}
& 0.1778 & 0.1543 & 2.2040
& 0.1623 & 0.1551 & 2.2120 \\
& 4 & \textbf{0.0631} & 0.0973 & \textbf{1.5360}
& 0.1184 & 0.1201 & 2.3440
& 0.0848 & 0.1124 & 2.1200 \\
& 5 & \textbf{0.0436} & 0.0688 & \textbf{1.6040}
& 0.0765 & 0.0943 & 2.1720
& 0.0706 & 0.0879 & 2.2240 \\

\bottomrule
\end{tabular}
\end{table*}

\subsection{Detailed Analysis of DTLZ1,2, and 7}
\begin{figure*}[t]
    \centering
    \subfloat[Average Utility]{
        \includegraphics[width=0.47\linewidth]{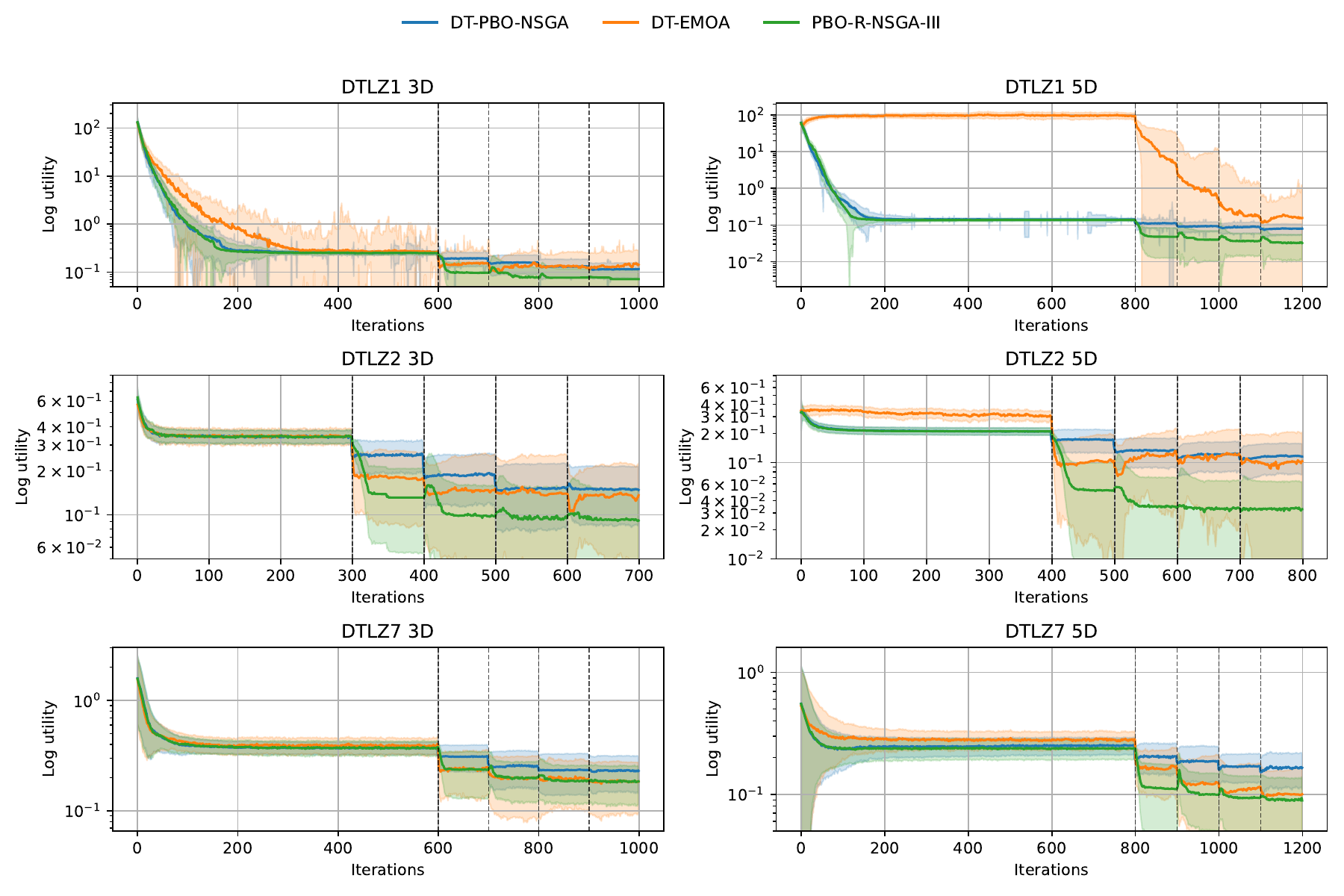}
        \label{fig:avg_utility_convergence}
    }
    \hfill
    \subfloat[Best-point Utility]{
        \includegraphics[width=0.47\linewidth]{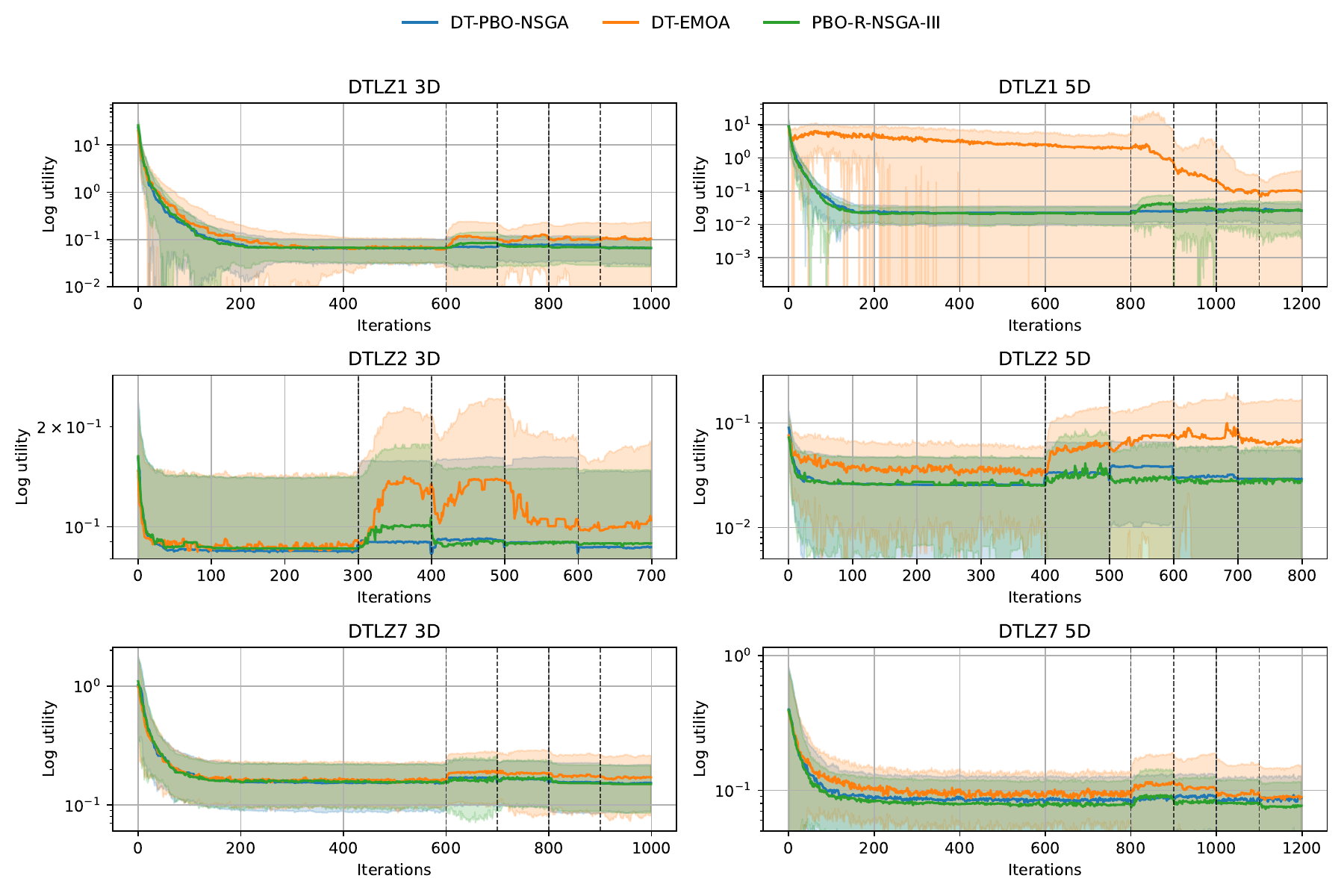}
        \label{fig:best_point_utility_convergence}
    }
    \caption{Convergence plots for 3- and 5-dimensional DTLZ1, 2, and 7. Subfigure (a) shows the median average utility of the final population and subfigure (b) shows the median best-point utility of the final population. Results are averaged over 50 trials. Confidence intervals show one standard deviation.}
    \label{fig:utility_convergence}
\end{figure*}

As the benchmark suite is too large to provide a detailed analysis for every individual problem, we focus on a small diverse subset. Specifically, we select DTLZ1, 2, and 7, as these problems have different Pareto front geometries. They are linear and multimodal, concave, and disconnected respectively. Moreover, for these results we only use Chebyshev generated artifical DMs. Figure~\ref{fig:avg_utility_convergence} and Figure~\ref{fig:best_point_utility_convergence} show convergence behavior by plotting the average utility and best-point utility over the amount of iterations ran. One can see that before any preference information is included DT-EMOA's base optimizer, NSGA-II, struggles with converging towards the Pareto front. This is caused by all solutions becoming non-dominated in higher dimensions. After introducing preference information, selection pressure is reintroduced and DT-EMOA's performance improves. However, for DTLZ1 it remains far off the other methods. 

The average utility convergence plots nicely illustrate what happens after preference information is introduced. Shortly after each preference round (denoted by the dashed vertical black lines), the average utility sharply drops as the model converges to the identified ROI. One can also see that those drops are usually smaller for DT-PBO-NSGA, indicating that the identified ROI is still larger than that of DT-EMOA and PBO-R-NSGA-III. The best-point utility plots, on the other hand, show that the best-point utility can actually increase after a preference elicitation round. This occurs when the ROI does not contain the most preferred solution.

\begin{figure*}
    \centering
    \includegraphics[width=\linewidth]{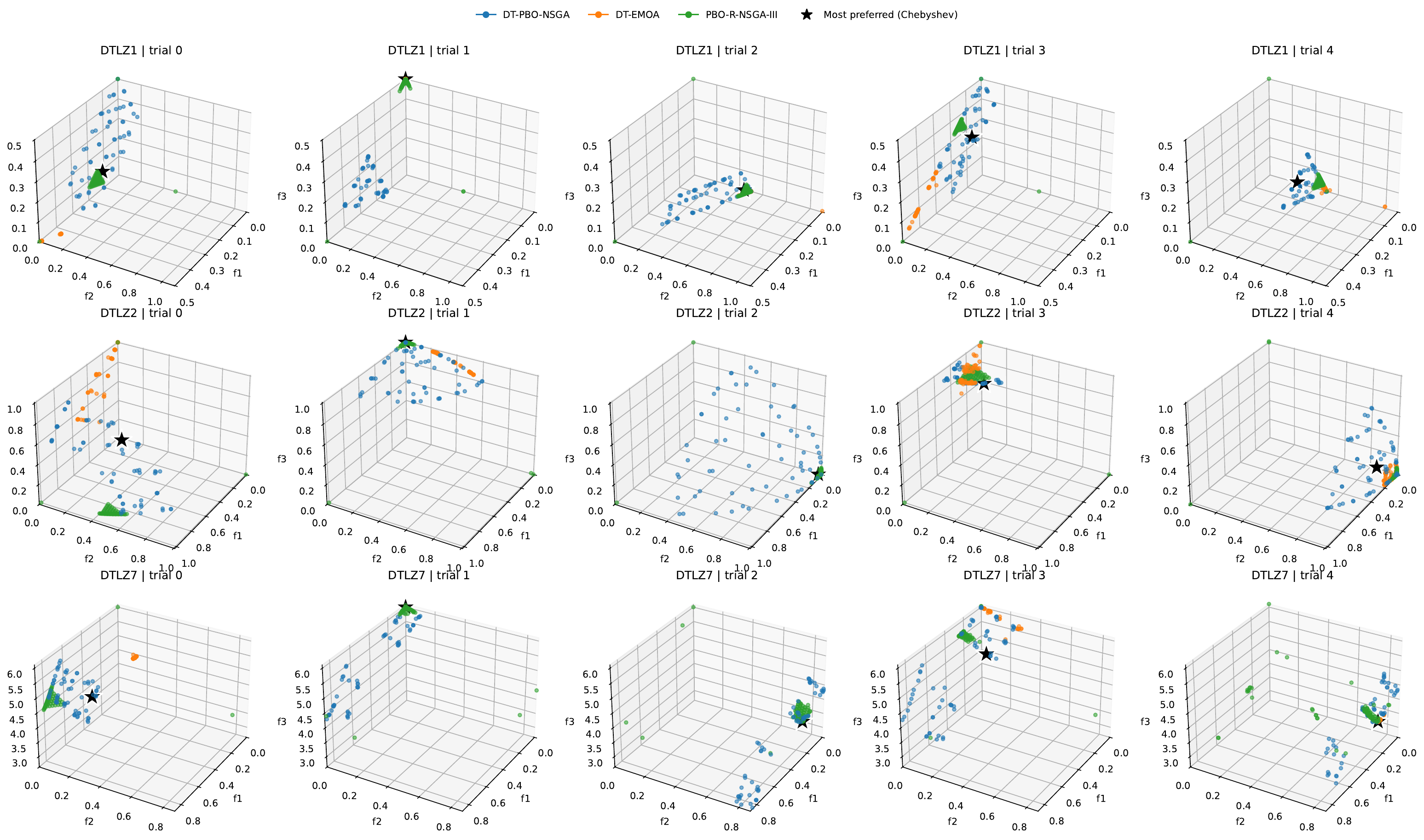}
    \caption{Final population plot for 3-dimensional DTLZ1, 2, and 7. }
    \label{fig:final_pop_3D}
\end{figure*}

\begin{figure*}
    \centering
    \includegraphics[width=\linewidth]{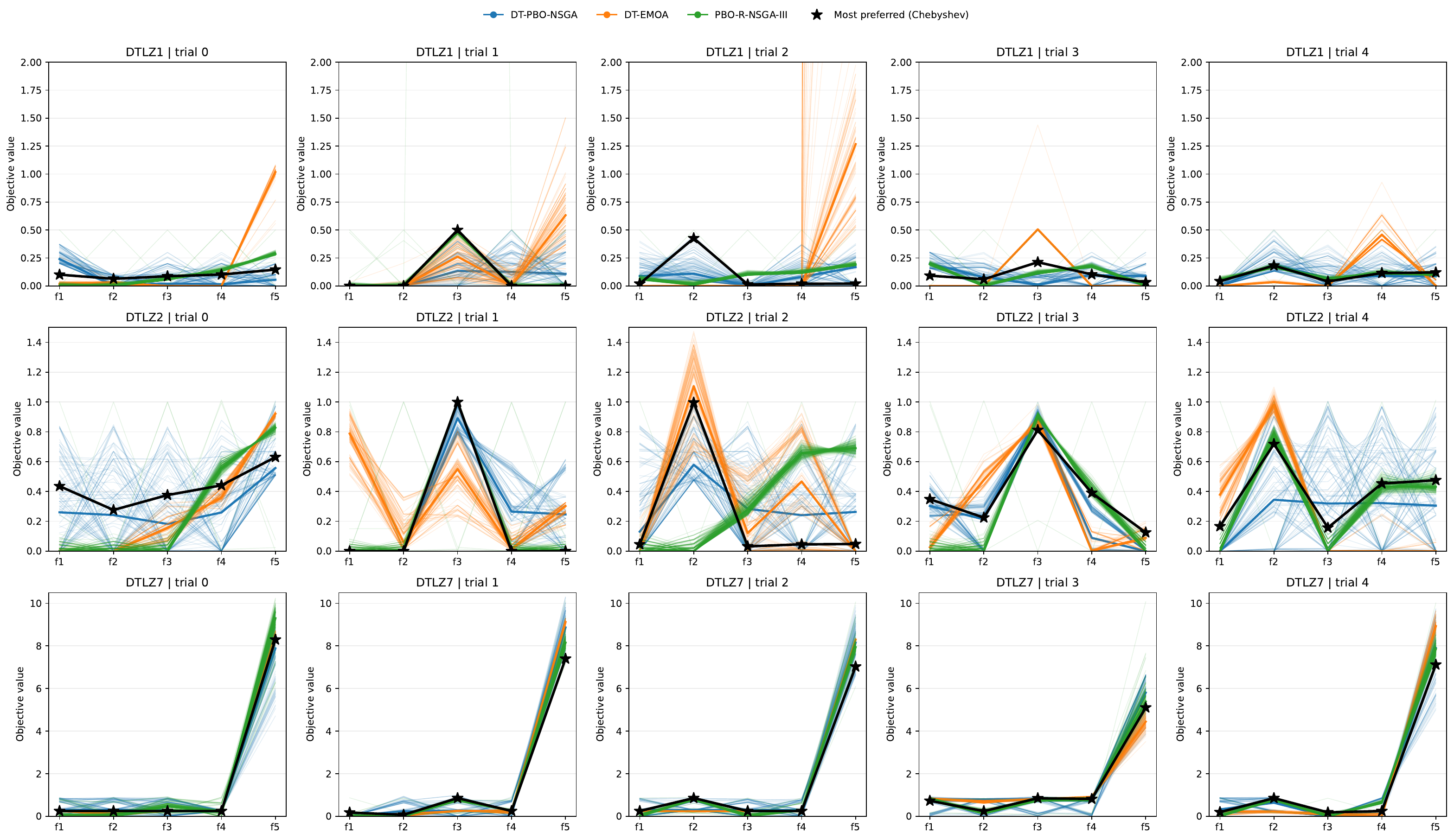}
    \caption{Final population plot for 5-dimensional DTLZ1, 2, and 7. For DTLZ1 trial 2, DT-EMOA's final population converges towards $f5=185$. For legibility the axis limit is set to $2$.}
    \label{fig:final_pop_5D}
\end{figure*}

The final population plots for the three-objective problems shown in Figure~\ref{fig:final_pop_3D} show the same trend observed before. DT-EMOA and PBO-R-NSGA-III find very small ROIs which reduce the average utility substantially. However, often the small ROIs do not actually contain the best point itself. For example, DTLZ 2 trial 0 both PBO-R-NSGA-III and DT-EMOA find ROIs far away from the most preferred solution. DT-PBO on the other hand maintains a much larger ROI but more often contains the best point. 5-dimensional population plots are included in Figure~\ref{fig:final_pop_5D}. These plots generally show the same trend, but also have some noteworthy differences. First, for DTLZ1 trial 2, DT-EMOA's final population diverges away from the Pareto front. This is likely caused by NSGA-II's poor performance in high-dimensional settings. For DTLZ2 trial 2 one can see that PBO-R-NSGA-III has a very concentrated ROI but it is very far from the most preferred solution. DT-PBO-NSGA, on the other hand, has a much broader ROI but does include the Most preferred solution.

\section{Utility Functions Visualized}

\begin{figure*}
    \centering
    \includegraphics[width=\linewidth]{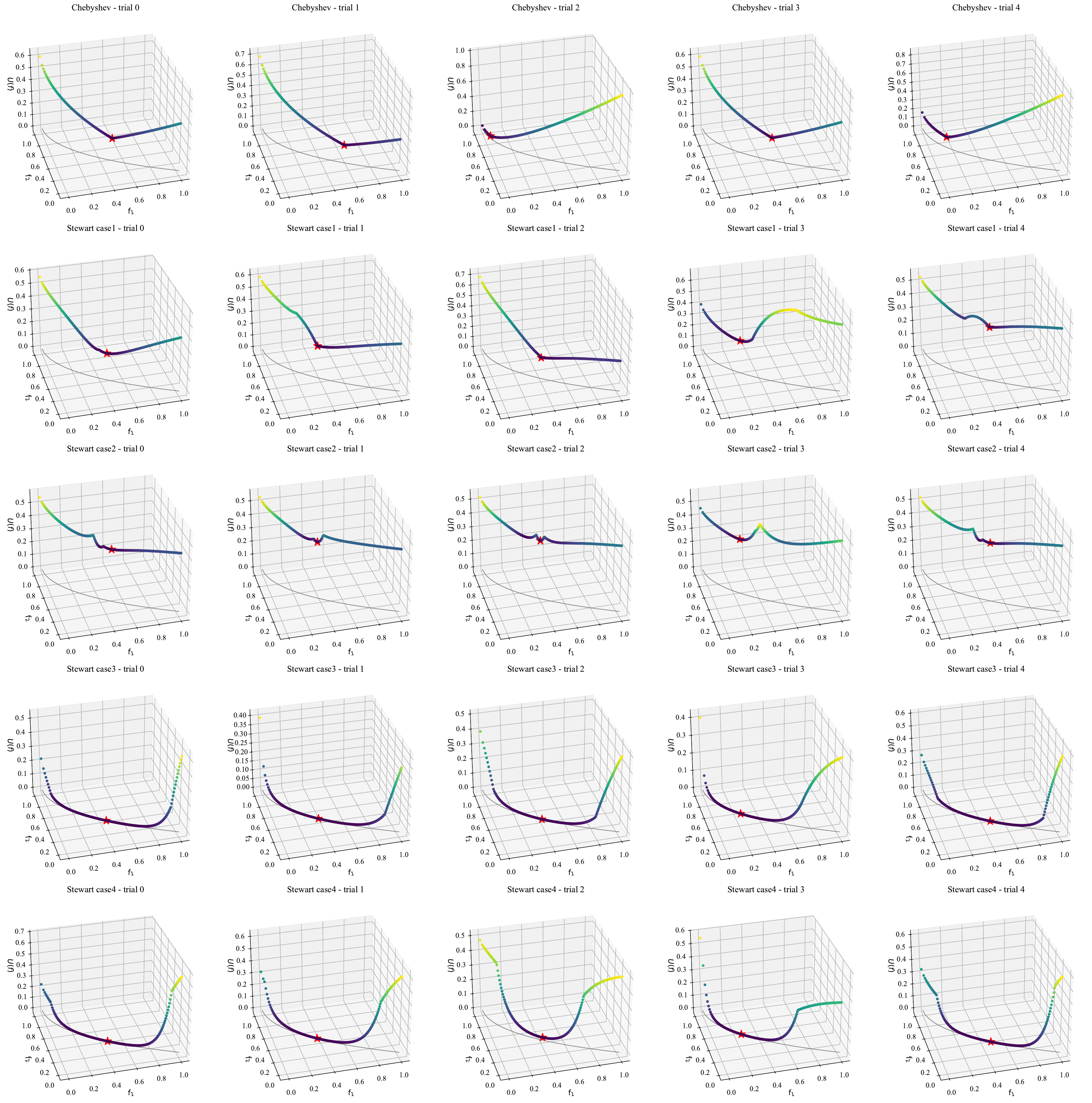}
    \caption{Utility values along ZDT1's Pareto front for 5 randomly generated DMs per utility function type.}
    \label{fig:uf_vis}
\end{figure*}

In Figure~\ref{fig:uf_vis}, we visualize the utility values along the Pareto front of ZDT1 for five randomly generated instances of each utility function type.  The Chebyshev utility is piecewise linear, with a single kink at the most preferred point. Stewart cases 1 and 2 vary more gradually along the front, although case 2 shows sharp local steps close to the optimum. Stewart cases 3 and 4, in contrast, exhibit long, flat valleys of near-optimal utility bounded by steep increases towards the extremes of the front, which level off near the boundary in some instances. For these cases, the effect of a change in objective values on utility therefore depends strongly on the position of a solution along the front.
\end{document}